\documentclass{article}
\usepackage{amsmath} 
\usepackage{hyperref} 
\usepackage{cleveref} 
\usepackage{algorithm}
\usepackage{amsmath}
\usepackage{amsthm}
\usepackage{amssymb}
\usepackage[framemethod=TikZ]{mdframed}
\usepackage{tikz}
\usepackage{pifont}
\usetikzlibrary{shapes.geometric, arrows.meta, positioning}

\tikzset{
  block/.style = {rectangle, draw, rounded corners, 
                  minimum width=3cm, minimum height=1cm, 
                  align=center, font=\small},
  arrow/.style = {-{Latex[length=3mm]}, thick}
}
\usepackage[font=small]{caption} 

\usepackage{algorithmic}
\usepackage{natbib}
\usepackage{mathtools}
\usepackage{amsfonts}
\usepackage{multirow} 
\usepackage{authblk}
\usepackage{xcolor}
\usepackage{subcaption}
\usepackage{subcaption}
\usepackage{booktabs}
\usepackage{multirow}
\usepackage{amsfonts}
\usepackage{caption} 
\usepackage{appendix}
\usepackage[utf8]{inputenc}
\usepackage{graphicx}
\usepackage{subcaption} 
\usepackage{bm}
\title{Causality-informed Anomaly Detection in Partially Observable Sensor Networks: Moving beyond Correlations}
\author[1]{Xiaofeng Xiao}
\author[2]{Bo Shen}
\author[1]{Xubo Yue\thanks{Corresponding Author: x.yue@northeastern.edu}}
\affil[1]{Mechanical and Industrial Engineering, Northeastern University}
\affil[2]{Mechanical and Industrial Engineering, New Jersey Institute of Technology}

\begin{document}

\maketitle

\begin{abstract}

Nowadays, as AI-driven manufacturing becomes increasingly popular, the volume of data streams requiring real-time monitoring continues to grow. However, due to limited resources, it is impractical to place sensors at every location to detect unexpected shifts. Therefore, it is necessary to develop an optimal sensor placement strategy that enables partial observability of the system while detecting anomalies as quickly as possible. Numerous approaches have been proposed to address this challenge; however, most existing methods consider only variable \textbf{correlations} and neglect a crucial factor: \textbf{Causality}. Moreover, although a few techniques incorporate causal analysis, they rely on interventions—artificially creating anomalies—to identify causal effects, which is impractical and might lead to catastrophic losses.

In this paper, we introduce a causality-informed deep $Q$-network (Causal DQ) approach for partially observable sensor placement in anomaly detection. By integrating causal information at each stage of $Q$-network training, our method achieves faster convergence and tighter theoretical error bounds. Furthermore, the trained causal-informed $Q$-network significantly reduces the detection time for anomalies under various settings, demonstrating its effectiveness for sensor placement in large-scale, real-world data streams. Beyond the current implementation, our technique's fundamental insights can be applied to various reinforcement learning problems, opening up new possibilities for real-world causality-informed machine learning methods in engineering applications.


\end{abstract}


\section{Introduction}


In modern complex networked systems, ranging from cyber-physical infrastructures and wireless sensor networks to industrial Internet-of-Things (IIoT) platforms, timely detection of anomalous behavior is critical for ensuring reliability, security, and efficient operation. Anomalies may arise from system faults, malicious attacks, unexpected environmental changes, or sensor malfunctions, often leading to degraded performance or catastrophic failures if left undetected. Consequently, anomaly detection has received significant attention from both the research community and industry practitioners. To cope with the challenges posed by increasingly large-scale, dynamic, and partially observable networks, there has been growing interest in developing anomaly detection techniques that can operate effectively under limited or incomplete information. One prominent line of anomaly detection research focuses on exploiting statistical dependencies and correlations among observed variables. Machine learning approaches, in particular, often leverage these relationships to distinguish normal patterns from anomalous behavior (See Section \ref{sec:LR} for a brief literature review).

However, considering only the correlations among variables is insufficient to fully exploit the potential of sensor placement strategies. A critical yet often overlooked factor is the underlying \textbf{causality} among variables. Causality describes a cause-and-effect relationship between variables. For example, in dataset $X$, if variable $X_i$ causes $X_j$, then changing $X_i$ will lead to a change in $X_j$. This is different from correlation, which only means $X_i$ and $X_j$ move together — not necessarily that one causes the other. By understanding the causality, we can infer hidden mean shift and detect anomalies earlier. Specifically, if variable $X_i$ causally influences variable $X_j$, then an anomaly occurring in $X_i$ will also induce an anomaly in $X_j$. In this case, placing a sensor on $X_i$ alone would be sufficient to detect mean shifts in both variables.

Although several recent studies \citep{huang2024causal, gao2024causal, succar2025detecting} have begun to incorporate causality into anomaly detection frameworks, uncovering causal relationships among variables remains challenging—particularly in settings with partially observable sensors. Moreover, identifying the true causal structure often requires interventions, which are difficult or even infeasible to perform in many real-world manufacturing environments. For example, under partial sensor placement, interpreting ``intervention" as the deliberate induction of anomalies would require intentionally introducing faults or malfunctions into equipment—an approach that is neither practical nor safe in real-world manufacturing environments. Therefore, it is essential to develop approaches that can infer optimal partial sensor placements based on the causality, without relying on interventions. Meanwhile, anomaly detection typically requires sequential and timely decision-making to accurately localize the variables exhibiting mean shifts. As such, the Partially Observable Multi-Sensor Sequential Change Detection (POMSCD) problem can naturally be formulated as a Partially Observable Markov Decision Process (POMDP). 

This paper proposes a causality-informed anomaly detection approach integrated with a $Q$-network. This causality-informed deep $Q$-network (Causal DQ) is essential for finding the optimal sensor placement and detecting mean-shifted data streams in a timely manner. The training workflow of the proposed Causal DQ framework consists of the following phases (see details of Figure~\ref{fig:workflow}):
\textbf{(1)} Causal discovery methods are first employed to estimate the causal relationships (i.e., the causal graph) among variables in the dataset. The resulting causal structure is then incorporated-along with other relevant factors into the construction of the system state, forming what we refer to as the \textbf{causal state}.
\textbf{(2)} Given the constructed \textbf{causal state}, the $Q$-network is trained to sequentially select variables with potential mean shifts as actions, corresponding to sensor placements. During this process, selecting a truly mean shifted variable is treated as a rewarding action, following the principle of ``action causes reward". The causal link between actions and observed rewards is encoded as a \textbf{causal mask}, which is used to compute a \textbf{causal entropy} term. This term is incorporated into the $Q$-network's loss function to guide training and encourage causally grounded exploration.
\textbf{(3)}  Through iterative updates, the $Q$-network regularized by causal entropy-converges to a policy that identifies optimal sensor placement strategies, which enables efficient and causally aware anomaly detection under partial observability. 

The contributions of this paper are described below: 
\begin{itemize}
    \item We propose scalable \textbf{Causal DQ}, a causality-informed reinforcement learning framework for partially observable sensor placement. Unlike existing causal RL approaches, our method is intervention-free—it avoids risky or invasive manipulations of system variables. This makes \textbf{Causal DQ} both practically feasible and scalable for real-world manufacturing applications. Therefore, this work addresses a critical research gap by introducing a \textbf{non-intervention, causality-informed} approach to anomaly detection under \textbf{partially observable} sensor configurations.

    \item \textbf{Causal DQ} incorporates causal component at every stage of the RL framework — not only in modeling the relationships among variables, but also in the reward design and updating procedures. This comprehensive, causality-informed framework enables our approach to be broadly scalable and adaptable across various RL paradigms.

    \item We provide rigorous mathematical proofs and derivations for the proposed \textbf{Causal DQ} framework, demonstrating that our causality-informed reinforcement learning approach outperforms its non-causal counterpart. Specifically, we show that \textbf{Causal DQ} satisfies the \textbf{contraction property, achieves tighter error bounds, and enjoys a faster convergence rate}.
\end{itemize}

Table~\ref{tab:comparison} presents a comparative analysis of our proposed \textbf{Causal DQ} framework against existing causality-informed or reinforcement learning-based anomaly detection approaches \citep{ zhang2023bandit, guo2024thompson, gao2024causal, huang2024causal,li2025online}.

\begin{table}[H]
\centering
\captionsetup{font=footnotesize}

\resizebox{\textwidth}{!}{
\begin{tabular}{@{}lcccccc@{}}
\toprule
\textbf{Method} & \textbf{Causality} & \textbf{Scalable to other RL} & \textbf{Math Foundations}  & \textbf{Intervention-Free}&\textbf{Correlation} & \textbf{Partial Observability} \\ \midrule
\textbf{Causal DQ} & $\checkmark$ & $\checkmark$& $\checkmark$ & $\checkmark$ & $\checkmark$ & $\checkmark$\\
\cite{li2025online} & \ding{55}& \ding{55} & \ding{55} & \ding{55} & $\checkmark$ & $\checkmark$ \\
\cite{zhang2023bandit} & \ding{55} &\ding{55} & $\checkmark$ &  \ding{55}& $\checkmark$ & $\checkmark$\\
\cite{guo2024thompson} & \ding{55} & \ding{55} & $\checkmark$ & \ding{55}& $\checkmark$ &$\checkmark$\\
\cite{gao2024causal} & $\checkmark$ & \ding{55} & $\checkmark$ & \ding{55} & $\checkmark$ & \ding{55}\\
\cite{huang2024causal} & $\checkmark$ & \ding{55} & $\checkmark$ &  \ding{55} & $\checkmark$ & \ding{55}\\
\bottomrule
\end{tabular}
}

\caption{Comparison of different causality-informed or reinforcement learning-based anomaly detection approaches. As shown, our proposed \textbf{Causal DQ} uniquely integrates \textbf{causality}, operates in an \textbf{intervention-free} manner, and is \textbf{scalable to general RL frameworks} with solid \textbf{mathematical foundations}. Unlike prior works, Causal DQ simultaneously addresses \textbf{correlation structure} and \textbf{partial observability}, highlighting its broad applicability and theoretical robustness.}
\label{tab:comparison}
\end{table}

We organize the remainder of this paper as follows: In Section~\ref{sec:LR}, we review related work on causal reinforcement learning and anomaly detection under partially observable sensors. In Section~\ref{sec:Presetting}, we introduce the problem formulation and preliminary definitions.
Section~\ref{sec:methodology} presents the workflow and algorithmic details of the proposed \textbf{Causal DQ} framework. The mathematical foundations and theoretical analysis are provided in Section~\ref{sec:maththeory}. In Section~\ref{sec:Experiment}, we report the results of extensive simulation experiments. Section~\ref{sec:realcase} demonstrates the effectiveness of our approach in a real-world case study. Finally, Section~\ref{sec:discuss} concludes the paper and outlines potential future directions.

\section{Literature Review}\label{sec:LR}

In this section, we provide a brief literature review on related work.

\paragraph{Correlation-based Partially Observable Sensors Network.} Due to resource constraints, it is often infeasible to monitor every data stream with sensors. This motivates extensive investigation into optimal sensor placement strategies for anomaly detection under partial observability \citep{chong2009partially, viti2014assessing}. Several recent studies have addressed optimal partial sensor placement. To name a few, \citet{shao2021partially} adopted the concept of minimum‐age scheduling to select the optimal subset of sensors.  \citet{lauri2022partially} investigated strategies for placing partial sensors in robotic systems, outlining methods to monitor robotic actions and movements efficiently.  \citet{dabush2023state} formulated sensor placement as a graphical model and derive state vectors that optimize deployment. Recognizing that anomaly detection under partial observability constitutes partially observable Markov decision processes (POMDPs), \citet{liu2022partially} demonstrated the feasibility of reinforcement learning (RL) techniques for learning optimal sensor placement policies. Meanwhile, \citet{liu2022sample} and \citet{kurniawati2022partially} apply RL under partial observability to multi‐agent and robotic systems, displaying  potential of RL for wider application of optimal sensor placement.
Moreover, \citet{yang2022optimal} and \citet{li2025online} extended standard reinforcement learning to deep RL (as $Q$‐learning) under partial observability by employing $Q$‐networks, thereby enabling scalable monitoring of high‐dimensional data streams. However, all of the above studies focus solely on statistical correlations among variables during monitoring and overlook causal relationships.

\paragraph{Causality-informed Machine Learning.} 
Some early efforts have explored causality in manufacturing across various scenarios \citep{li2007knowledge, li2008causation, xian2018causation}, such as control process or data diagnosis. More recently, causality-informed machine learning has attracted increasing attention across a wide range of fields \citep{li2024causality, guo2024causality}. Causal information can significantly improve machine learning models by capturing how variables affect each other.  A typical ML approach investigates associations in a dataset, but mere associations between variables cannot account for all the variation in the data \citep{kaddour2022causal}. For example, in the design of electrohydrodynamic (EHD) printers, we might observe a positive relationship between temperature and the width of the printed line \citep{xiao2025explainable}. However, this does not mean that we can widen the line simply by raising the temperature. In fact, the causal root is the applied voltage: increasing voltage both raises temperature and broadens the line. \cite{cheng2022evaluation} highlighted the potential of causal-ML techniques in computer vision applications, enabling the identification of the underlying causes of algorithmic misinterpretations in image analysis. Furthermore, \cite{naser2022causality} discussed how causal ML methods can be applied in civil and structural engineering to uncover the many hidden covariates that affect system behavior. 
Moreover, \cite{lechner2023causal} showed how causal-ML methods helped in social sciences, for example by quantifying economic factors, and \cite{feuerriegel2024causal} discussed how causal ML makes medical treatment outcome prediction more robust when the underlying causal relationships were known. 

Despite its promise, very few studies have explored causality-informed anomaly detection. \cite{tang2024causal} proposed a causal counterfactual graph neural network for anomaly detection in healthcare. This approach identifies features affected by sampling bias, thereby enhancing the clinical reliability of Parkinson’s disease diagnosis. Similarly, \cite{li2024causality} developed a causality-informed neural network to predict and detect anomalous features in CT images, leading to improved diagnostic performance for pancreatic cancer. Most recently, \cite{liu2025crcl} proposed a Causal Representation Consistency Learning (CRCL) approach to model video normality, which enables both the detection and suppression of anomalies during video generation. Moreover, \cite{xing2025multi} presented a causal neural network for multi-dimensional anomaly detection in microservice architectures. This approach can also be applied to identify anomalous signals in intelligent sensing systems.


\paragraph{Causal Reinforcement Learning.} As causal discovery and inference have recently attracted significant attention across fields such as machine learning, epidemiology, and manufacturing, researchers have begun integrating causal information into reinforcement learning to enhance its potential and capabilities \citep{zhu2019causal, li2021causal, deng2023causal, zeng2024survey}.  \citet{seitzer2021causal} incorporated causal dependency structures among variables to improve exploration efficiency and off‐policy learning.  \citet{gasse2021causal} studied how the causal effects of interventions on variables influence the RL training process, explicitly addressing confounders during training to further boost learning efficiency. Recently, \citet{ruan2023causal} addressed unobserved confounders in imitation learning by combining causal discovery into the training process.  Similarly, \citet{ji2024ace} and \citet{cao2025causal} applied causal reinforcement learning to robotic systems, enabling algorithms to capture and leverage the causal relationships among actions, states, and rewards. \textbf{Notably, most of the aforementioned methods require interventions on system variables to infer causal relationships, and such interventions may be infeasible or even detrimental in real‐world scenarios.}

 
As anomaly detection always requires timely, sequential decision-making, reinforcement learning is a natural fit for partial sensor placement in this context \citep{oh2019sequential,  zha2020meta, arshad2022deep}, and the causal information among data streams encodes how mean shifts in one variable propagate to other variables that are causally connected to it.

However, only a few studies have explored integrating causal reinforcement learning into partially observable sensor placement for anomaly detection.  Especially, the key challenge is that existing causal RL methods \citep{grimbly2021causal, wang2021ordering, lu2022efficient, shi2023dynamic,liao2024instrumental} often \textbf{rely on interventions to infer causality among variable}s—a requirement that is infeasible and impractical in real‐world applications. For instance, in the context of partial sensor placement, treating an “intervention” as an induced anomaly implies deliberately causing equipment faults or malfunctions—an approach that is both impractical and potentially damaging.

Our proposed Causal DQ is an intervention‐free causal reinforcement learning framework for optimal sensor placement, effectively addressing the limitations of existing methods.







\section{Preliminary Setting}\label{sec:Presetting}



\subsection{Anomaly Detection with
Partially Observable Sensors Network}

We consider a $p$-dimensional dataset $X$, where $p\in\mathbb{Z}^+$ denotes the number of variables in the system. All data streams at each time epoch  $t\in\{1,\dots,n\}$ are represented by the vector $X(t) = \bigl[X_1(t), \ldots, X_p(t)\bigr]^\top \in \mathbb{R}^p$. The sequence $\{X(t)\}_{t=1}^n$ is distributed with density $P\bigl(X(t)\mid\mu\bigr)$, where $\mu\in\mathbb{R}^p$ is the mean vector of the distribution. Due to resource constraints, only $m$ out of the $p$ streams can be monitored at each epoch, where $m\le p$. This partially observed vector is defined as $X_m(t) = \bigl[X_j(t)\bigr]_{j\in S}
= [X_{j_1}(t), X_{j_2}(t), \dots, X_{j_m}(t)]^\top
\in \mathbb{R}^m$, where  $S \subseteq \{1,2,\ldots,p\}, |S|=m$, is the set of selected sensor indices.

At an unknown change point $T$, one or more variables experience a mean shift, so that observations at $t=T$ follow the shifted distribution $P\bigl(X(T)\mid\mu_s\bigr)$, where $\mu_s\neq\mu$ denotes the post-change mean. Without loss of generality, we set the pre-change mean $\mu=0$ and denote the nonzero shifted mean by $\mu_s\neq 0$. Since the number of affected streams is unknown and the total number of sensors is limited to $m$, we must develop a monitoring framework that both detects and localizes mean-shifted streams under partial observation. In this work, we adopt the standard statistic-based alarm-triggering approach to detect mean shift. Specifically, at each time $t$ we compute a detection statistic $\Lambda(t)$ from the $m$ monitored data streams $X_{m}(t)$. If $\Lambda(t)$ reaches or exceeds the predefined threshold, an alarm is raised.

Generally, two common metrics are used to evaluate the performance of monitoring schemes for anomaly detection under partial observation:

(1) \textbf{Average Run Length (ARL)}: the expected number of time steps from the \textbf{start of monitoring} until a false alarm is triggered when no change has occurred (i.e., under the null hypothesis).

(2) \textbf{Average Detection Delay (ADD)}: the expected number of time steps from the \textbf{occurrence of an anomaly (i.e., a mean shift)} until the detection scheme raises an alarm.

These two metrics quantify the trade-off between timely detection and false alarm control in the monitoring process. This paper employs the \textbf{Average Detection Delay (ADD)} metric to assess the performance of our proposed scheme.

\textit{However, traditional methods for partially observable sensors only consider correlations while neglecting an equally important factor: causality between data streams.}  Causality enables us to determine whether a mean shift in one stream will propagate and induce anomaly behavior in downstream variables. For example, if variable $X_i$ has a direct causal effect on $X_j$, then a mean shift in $X_i$ is likely to cause an mean shift in $X_j$, so we only need to place a single sensor at the location of  $X_i$ instead of monitoring both $X_i$ and $X_j$ simultaneously, reducing resource costs and improving monitoring efficiency. Therefore, considering causal relationships among data streams is essential when determining optimal sensor placement under partial observability.   

\subsection{Causal Discovery and Causal Component Effect}\label{sec:3.2causaldiscover}

The causal discovery methods (e.g., the Peter-Clark (PC) algorithm and Fast Causal Inference (FCI) algorithm) aim to uncover causal relationships among a set of variables $X = \{X_1, \ldots, X_p\}$ \citep{glymour2019review}. First, a complete undirected graph—known as the skeleton—is constructed, where each node represents a variable and every pair of nodes is initially connected by an edge. Conditional independence tests are then performed on this skeleton: for any pair $(X_i, X_j)$, if $X_i$ and $X_j$ are found to be conditionally independent given some conditioning set $S \subseteq X \setminus \{X_i, X_j\}$, the edge between them is removed \citep{kalisch2007estimating, entner2010causal, harris2013pc}. Finally, without introducing cycles or new unshielded colliders, as many of the remaining edges as possible are oriented, resulting in a Completed Partially Directed Acyclic Graph (CPDAG) that represents the equivalence class of Directed Acyclic Graphs (DAG) consistent with the observed conditional independencies. 

In the DAG, each node $X_i$ corresponds to a variable, and a directed edge $X_i \to X_j$ indicates that $X_i$ is a direct cause (parent) of $X_j$, denoted $X_i \in \mathrm{PA}(X_j)$. Any indirect path $X_i \to \cdots \to X_j$ implies that $X_i$ is an indirect cause (ancestor) of $X_j$, written $X_i \in \mathrm{AN}(X_j)$, with $\mathrm{PA}(X_j)\subseteq \mathrm{AN}(X_j)$.  The strength of these causal links is quantified by the conditional distribution $P\bigl(X_j \mid \mathrm{PA}(X_j)\bigr)$ or $P\bigl(X_j \mid \mathrm{AN}(X_j)\bigr)$, which can be readily obtained by Bayesian Network (BN) \citep{pearl1995bayesian, korb2010bayesian, marcot2019advances,kitson2023survey, chen2024temporal}. Bayesian Networks exploit the DAG structure by factorizing the joint probability distribution over all variables into a product of local conditional distributions: $P(X_1, X_2, \dots, X_n) = \prod_{i=1}^n P\bigl(X_i \mid PA(X_i)\bigr).$
This factorization  makes explicit the causal relationships encoded in the network by their the joint distribution, as each conditional probability represents a causal mechanism between a variable and its direct causes.



\subsection{Reinforcement Learning in Partially Observable Anomaly Detection}\label{tab:RL Intro} 

Anomaly detection in resource‐constrained environments demands a sequential, timely strategy for dynamically placing sensors and periodically adjusting those placements based on changing environmental information. Consequently, reinforcement learning naturally lends itself to this problem, as it can learn an optimal sensor‐selection policy that balances detection speed with sensing cost. There are multiple approaches of RL have been adopted to do the Anomaly Detection with partial sensor placements. For instance, \cite{zhang2019partially} extended combinatorial multi‐armed bandits to identify subset of sensors that minimize the ADD under a sensing budget; \cite{li2024online} trained a $Q$‐learning to sequentially select sensor locations, aiming to reduce the ARL until detection. However, traditional RL methods struggle with high‐dimensional sensor placement, since the number of possible actions grows combinatorially—for example, selecting 5 sensors out of 100 yields $\binom{100}{5} = 75,287,520$ possible combinations! For $Q$-learning approach, it is infeasible to build a $Q$-table with $\binom{100}{5}$ columns to contain all possible actions. To address this intractable action space, deep RL techniques such as the deep $Q$‐network can be employed to approximate the optimal action‐value function and efficiently learn sensor‐selection policies \citep{hester2018deep, fan2020theoretical}. $Q$-network uses a neural network to approximate the mapping from states and actions to $Q$‑values directly, eliminating the need for an explicit table. Moreover, the network naturally generalizes to datasets of varying size then making the trained network more scalable.

However, a single $Q$-network can cause overestimation bias during training because the network may fall into suboptimal actions at an early stage, and other optimal actions can be neglected. To address this challenge, a typical approach is to adopt a Double $Q$-network (DQN) to separately select actions and compute the $Q$-values \citep{van2016deep,li2025online}. Specifically, we need two $Q$-networks, $Q_{online}$ and $Q_{target}$. During each training iteration, $Q_{online}$ is used to select the optimal action $a'$ at the updated state $s'$ as follows:
$$
a^* = \arg \max_{a'} Q_{online}(s', a'; \theta_{online})
$$
where $\theta_{online}$ denotes the parameters of $Q_{online}$. Once the optimal action $a^*$ is selected, we employ $Q_{target}$ together with the immediate reward $r$ to compute the accumulated $Q$-value $y$:

\begin{equation}\label{eq:accumulated-q-value}
y = r + \gamma \cdot Q_{target}(s', a^*; \theta_{target})
\end{equation} with $\theta_{target}$ being the parameters of $Q_{target}$, and $\gamma$ is decayed factor in updating. Finally, the temporal difference (TD) is computed by sampling transitions $(s, a, r, s')$ from the experience replay, where $s$ is the current state, $a$ is the action taken, $r$ is the reward received, and $s'$ is the updated state. This leads to the following loss function:
\begin{equation}\label{eq:loss-TD}
\mathcal{L}(\theta_{online}) = \mathbb{E}_{(s, a, r, s') \sim \text{replay}} \left[ \Bigl( y - Q_{online}(s, a; \theta_{online}) \Bigr)^2 \right].
\end{equation}

Although deep $Q$-networks can mitigate the high-dimensional action-space challenge in sensor placement, most existing methods consider only statistical correlations and ignore the underlying causality among data streams. To address this shortcoming, we propose a \textbf{Causal DQ} framework, which incorporates causal information into the action-value estimation to improve detection accuracy and robustness under partial observability.

\section{Methodology}\label{sec:methodology}
The \textbf{key contribution} of our work is that we incorporate causal information at every stage of training $Q$-network for anomaly detection under partial observability. Specifically, we modify the conventional Bellman objective by adding a causal‐entropy regularizer. Furthermore, we include causal information in each state by including the causal‐statistic components.  Moreover, we incorporate the causal information of each state–action pair directly into the design of the reward function, ensuring that rewards reflect causal impact from action. \textbf{To our knowledge, this is the first work that studies causality-informed anomaly detection under partial observability.}



\begin{figure}[H]
  \centering
  \includegraphics[width=1\textwidth]{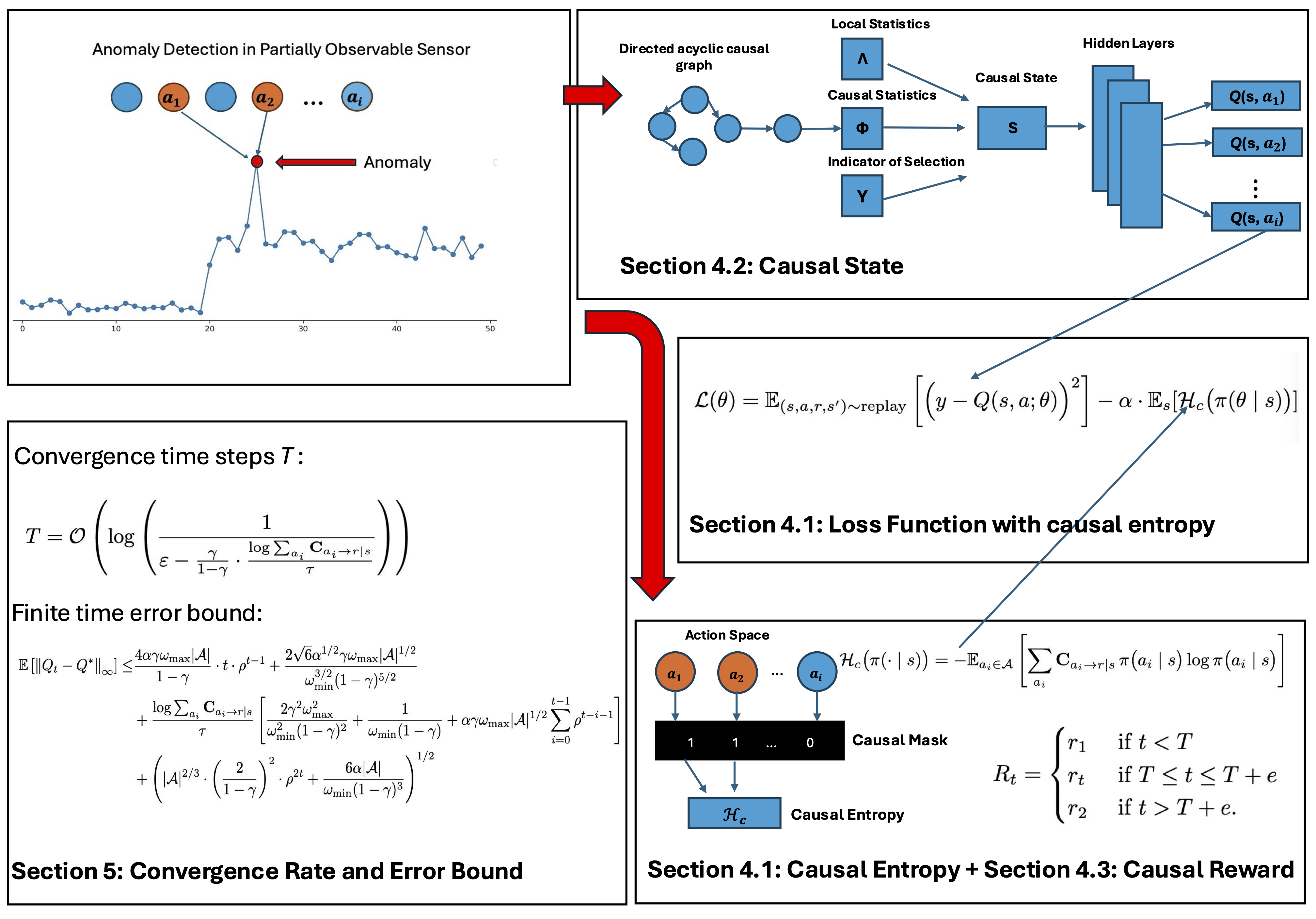} 
  \caption{Workflow of our proposed causality-informed anomaly detection under partial observability.} 
  \label{fig:workflow}
\end{figure}

We provide the details of our approach in this section (Fig. \ref{fig:workflow}). We extend the traditional deep $Q$-network to our proposed Causal DQ in Section \ref{tab:causalEntropy}. In Section \ref{sec:causalSA}, we describe the ideas and process to construct the ``causal statistic'', which is a critical element to be used in the \textbf{causal state} of our Causal DQ. Then, we outline the details of the state and action in the Causal DQ, and how the causal state and action transition in a Markov Decision Process. The causal reward is expressed in Section \ref{tab:reward}. In Section \ref{tab:Offline}, we illustrate how the Causal DQ selects variables exhibiting a mean shift and triggers an abnormal alarm offline. We note here that, although our focused RL model is DQ, our idea of causality-informed ML can be extended to broader RL domains.


\subsection{Strategy Selection and Causal Entropy}\label{tab:causalEntropy} 

First, we incorporate causality into the $Q$-value update process, ensuring that each learned sensor-placement policy embeds the causal relationships among actions, states, and rewards.

We adopt the Boltzmann softmax to randomly select an action in order to balance exploration and exploitation. The Boltzmann softmax transforms the $Q$-value of the select action $a$ from action space $\mathcal{A}$ under state $s$ into a probability by computing
$$
\pi(a \mid s)=\frac{\exp\bigl(\frac{Q(s, a)}{\tau}\bigr)}{\sum^{|\mathcal{A}|}_{i=1}\exp\bigl(\frac{Q(s, a_i)}{\tau}\bigr)}
$$
where $\left\{a_i\right\}_{i=1}^{|\mathcal{A}|}$ denotes the set of all possible actions given the state $s$, and $|\mathcal{A}|$ is the cardinality of the action space. $\tau$ is a temperature parameter that regulates the trade-off between exploration and exploitation. This probability distribution yields an entropy defined as
$$
\mathcal{H}\bigl(\pi(\cdot \mid s)\bigr)=-\sum^{|\mathcal{A}|}_{i=1} \pi(a_i \mid s)\log \pi(a_i \mid s).
$$
Based on this entropy, we employ a causal entropy by incorporating a causal mask $\mathbf{C}_{a_i\rightarrow r\mid s}$, defined as \citep{ji2024ace, cao2025causal}:
$$
\mathcal{H}_c\bigl(\pi(\cdot\mid s)\bigr)=-\mathbb{E}_{a_i\in \mathcal{A}}\left[\sum_{a_i}\mathbf{C}_{a_i\rightarrow r\mid s}\,\pi\bigl(a_i\mid s\bigr)\log \pi\bigl(a_i\mid s\bigr)\right].
$$

Here, $\mathbf{C}_{a_i\rightarrow r\mid s} \in \mathbb{R}^{m\times 1}$ is a vector that encodes the causal relationship from an action to the reward, conditioned on the state $s$, which is similar to the causal mask $M$ employed in the reward function (see Section \ref{tab:reward}). In the scenario of this work, selecting a data stream with a mean shift results in a positive reward. To simplify the computational complexity and streamline the approach, we utilize a binary vector to represent $\mathbf{C}_{a_i\rightarrow r\mid s}$, assigning a value of $1$ to an action $a_i$ if it selects a data stream with a mean shift, and $0$ otherwise, which can be defined as: $\mathbf{C}(a,s): \mathcal{A} \times \mathcal{S} \rightarrow\{0,1\}$ with action space $\mathcal{A}$ and state space $\mathcal{S}$. In this way, each element simply indicates that choosing a data stream with a mean shift has a positive causal effect for that region. This causal entropy provides an effective regularization term for the $Q$-function, ensuring that its updates do not converge to sub-optimal solutions and thus enhancing the network's stability and scalability.

Moreover, the causal entropy $\mathcal{H}_c$ is well-defined and effective in both discrete and continuous action spaces. As $\pi(a \mid s)$ is a probability measure, the term $\mathbf{C}(a,s)\pi(a \mid s)$ can be defined as a new probability measure: $$P_c(a \mid s)=\frac{\mathbf{C}(a, s) \pi(a \mid s)}{\int_{\mathcal{A}} \mathbf{C}\left(\tilde{a}, s\right) \pi\left(\tilde{a} \mid s\right) d \tilde{a}}$$ over the continuous action space  $\mathcal{A}$, where $\tilde{a}$ is a continuous action variable integrated over the action space $\mathcal{A}$, which can gives out the causal entropy as: $$\mathcal{H}_c\bigl(\pi(\cdot\mid s)\bigr)=-\int_{\mathcal{A}} \log \pi(a \mid s) \mathrm{d} P_c(a \mid s).$$


As a result, by adopting the causal entropy, Eq.~\eqref{eq:accumulated-q-value} associated with the causal entropy is represented as:
\begin{equation}\label{eq:entorpy-Q}
y = r + \gamma \cdot Q_{target}(s', a^*; \theta_{target}) + \mathcal{H}_c\bigl(\pi(\cdot\mid s)\bigr).
\end{equation}


Meanwhile, Eq.~\eqref{eq:loss-TD} associated with causal entropy is expressed as:

\begin{equation}\label{eq:entorpy-TD}
\mathcal{L}(\theta_{online}) = \mathbb{E}_{(s, a, r, s') \sim \text{replay}} \left[ \Bigl( y - Q_{online}(s, a; \theta_{online}) \Bigr)^2 \right] - \alpha \cdot \mathbb{E}_s[\mathcal{H}_c\bigl(\pi(\theta_{online}\mid s)\bigr)].
\end{equation}


To learn the optimal online parameters $\theta_{online}$ for $Q$-network function, we minimize the total loss function by computing its gradient $\nabla_{\theta_{online}} \mathcal{L}$. Especially, the gradient of the causal entropy $\mathcal{H}_c$ under parameter $\theta$ is: 
\begin{align*}
\nabla_{\theta} \mathcal{H}_c &= - \sum_{a_i} \mathbf{C}_{a_i \rightarrow r | s} \cdot \left( \log \pi(a_i \mid s) + 1 \right) \cdot \nabla_{\theta} \pi(a_i \mid s) \\
 &= - \sum_{a_i} \mathbf{C}_{a_i \rightarrow r | s} \cdot \left( \log \pi(a_i \mid s) + 1 \right) \cdot \pi(a_i \mid s) \cdot \left( \frac{\nabla_{\theta} Q(s, a_i)}{\tau} - \mathbb{E}_{a'} \left[ \frac{\nabla_{\theta} Q(s, a')}{\tau} \right] \right) \\
&= -\mathbb{E} \left[ \sum_{a_i} \mathbf{C}_{a_i \rightarrow r | s} \cdot \pi(a_i \mid s) \left( \log \pi(a_i \mid s) + 1 \right) \cdot \frac{\nabla_{\theta} Q(s, a_i)}{\tau} \right].
\end{align*}

Substitute the gradient of $\mathcal{H}_c$ into the $\nabla_{\theta_{online}} \mathcal{L}$ , and then this overall gradient is: 
$$\nabla_{\theta_{online}}\mathcal{L}=-2 \mathbb{E}\left[\left(y-Q_{\text {online }}\right) \nabla_\theta Q_{\text {online }}\right]+\alpha \mathbb{E}_s\left[\sum_{a_i} \mathbf{C}_{a_i \rightarrow r \mid s} \cdot \pi(a_i \mid s)(\log \pi(a_i \mid s)+1) \frac{\nabla_{\theta_{online}} Q}{\tau}\right].$$

It can be seen that the proposed $Q$-value function is regularized by causal entropy at each update step. This causal entropy term, $\mathcal{H}_c$, encourages the network to maintain a level of uncertainty or randomness in its action selection, thereby mitigating overfitting to spurious or transient correlations in the data. Importantly, the causal mask $\sum_{a_i}\mathbf{C}_{a_i \rightarrow r \mid s}$ enhances the $Q$-function's sensitivity to the causal relationship between actions and rewards. This property is particularly advantageous in high-dimensional or partially observed environments, such as those encountered in industrial monitoring scenarios. By incorporating this causal entropy, the resulting $Q$-network benefits from tighter error bounds and faster convergence. The mathematical derivations and theoretical guarantees are provided in Section~\ref{sec:maththeory}.


\subsection{Causal State and Action}\label{sec:causalSA}

In addition to integrating causality into the $Q$-value updates, we capture the causal relationships among variables by introducing a \textbf{causal statistic} that quantifies the influence of causality across data streams. Moreover, we maintain a \textbf{local statistic} that is updated recursively over time, and an alarm is triggered once it exceeds a predefined threshold. To track the observable data streams at each time step, we also define an \textbf{indicator of selection}, which identifies the selected subset of variables.

These three components — the Local Statistic, Causal Statistic, and Indicator of Selection — collectively define the \textbf{Causal State} in this work.



\paragraph{Local Statistic} For a data stream $X(t)=\left[X_1(t), \ldots, X_p(t)\right]$ with $p$ dimensions at time $t$, $t = 1, \ldots, n$, we assume that the data stream follows a Gaussian distribution $p(X(t) \mid \mu)$ with covariance matrix $\Sigma$ , where $\mu$ is the mean vector. 

As all the data streams are observed over time, the distribution is recursively updated as  
$$
p\left(\mu \mid X(t), \ldots, X(n)\right) = 
p\left(\mu \mid X(1), \ldots, X(n-1)\right)^{1-\lambda} 
p\left(X(n) \mid \mu\right)
$$  
where $\lambda$ is the time-decay parameter \citep{zhang2019partially}. The covariance matrix is incrementally updated as  
$
V_n^{-1} = (1-\lambda) V_{n-1}^{-1} + \mathbf{E}^{T} \Sigma(n)^{-1} \mathbf{E},
$  
and the mean vector is updated similarly:  
$
\mu_n = V_n \left[(1-\lambda) \mu_{n-1} + \mathbf{E}^{T} \Sigma(n)^{-1} X(n)\right],
$  
where $\mathbf{E} \in \mathbb{R}^{p \times p}$ is an identity matrix. The resulting local statistic is defined as $\Lambda(n) = \left(0 - \mu_n\right)^{T} V_n^{-1} \left(\mathbf{0} - \mu_n\right) = \mu_n^{T} V_n^{-1} \mu_n$.

Alternatively, this statistic can be expanded as $$\Lambda(n) = \mu_n^T V_n^{-1} \mu_n = \sum_{i=1}^p \sum_{j=1}^p \mu_{n,i} v_{ij} \mu_{n,j}$$ for total $p$ variables with each corresponding $i^{th}$, $j^{th}$ element in the updated mean vector and covariance matrix.

To incorporate this information into the state, we further investigate the contribution of each individual variable to this local statistic. First, we represent the latter two terms as a vector: $ \xi_n = V^{-1}_n \mu_n = [\xi_{n,1}, \xi_{n,2}, \ldots, \xi_{n,p}]^\top $, where $ \xi_{n,i} = \sum_{j=1}^p v_{ij} \mu_{n,j} $. Then, the local statistic can be written as: $ \Lambda(n) = \sum_{i=1}^p \mu_{n,i} \, \xi_{n,i} $, which implies that the contribution of each variable can be expressed as: $ \hat{\Lambda}_i(n) = \mu_{n,i} \xi_{n,i} = \mu_{n,i} \sum_{j=1}^p v_{ij} \mu_{n,j} $. This local statistic allows us to construct a confidence interval $\mathcal{CI}$ at the significance level $1 - \zeta$ based on the $\chi^2$ distribution with $p$ degrees of freedom. Specifically, the confidence region is defined as: $\mathcal{CI}=\{\mu\mid(\mu_n^T V_n^{-1} \mu_n \leq \chi_{p, 1-\zeta}^2\}$. 

This process can be interpreted as a general hypothesis testing procedure, where the null hypothesis $\mathbb{H}_0$ and the alternative hypothesis $\mathbb{H}_1$ are defined as follows. Under $\mathbb{H}_0$, the monitored data stream is assumed to be operating under normal conditions, implying that the true mean vector satisfies $\mu = 0$. In this case, the local test statistic $\Lambda(n)$ asymptotically follows a chi-squared distribution with $p$ degrees of freedom:
$$\Lambda(n) \xrightarrow{d} \chi_p^2.$$

The alternative hypothesis $\mathbb{H}_1$ corresponds to the case where $\mu \neq 0$, indicating a deviation from the nominal condition, which may suggest the presence of mean shift in the data stream.

Finally, if the sum of the statistics of all selected data streams, $ \Lambda(n) = \sum_{i=1}^m \hat{\Lambda}_i(n) $, exceeds the testing threshold $ \chi^2_p $, the abnormal alarm will be triggered.

\paragraph{Causal Statistic} Causal discovery methods are employed to estimate the causal propagation effect (CPE) $\eta_{ij}$ between variables $X_i$ and $X_j$, where $\eta_{i j}$ quantifies the strength of the direct causal effect of $X_i$ on $X_j$ under the estimated equivalence class. Specifically, the CPE $\eta_{ij}$ is defined as:  
\begin{equation*}
    \tilde{\eta}(X_i, X_j)
\begin{cases}
\in (0, 1) & \text{if } X_i \in \mathbf{AN}(X_j), \\
 =1 & \text{if } i = j, \\
 =0 & \text{otherwise}
\end{cases}
\end{equation*}
where $\mathbf{AN}(X_j)$ denotes the set of ancestor variables of $X_j$. In our work, any causal discovery algorithm is applicable.

Inspired by the construction of the local statistic $\Lambda(t)$, we follow a similar process to build this causal statistic. We define the incremental causal statistic of each variable itself as:  
$\Gamma_i(n) = \mu_{n, i}^2 \eta_{ii}$. Since the $\eta_{ii} = 1$, the self CPE of variable $i$ is always equals to its own updated $\mu^2_{n,i}$. To incorporate the causal propagation effects from all other variables, we define the total causal statistic for variable $i$ as:  
${\phi}_i = \mu_{n, i}^2 \eta_{ii} + \sum_{j \neq i} \mu_{n, i} \eta_{ij} \mu_{n, j}$. Finally, we can have a causal statistic vector $\mathbf{\Phi}(t) = \left[\phi_1, \ldots, \phi_p\right]$ associated to total $p$ variable. This statistic quantifies the extent to which a variable causally influences other covariates. For example, a larger value of ${\phi}_i$ indicates that changes in variable $i$ are likely to cause greater changes in other variables. This statistic helps enhance the state representation by incorporating causal information.

\paragraph{Indicator of Selection} We use an indicator $Y$ of state updating that helps the transition of state and action follows the Markov Decision Process \citep{li2025online}. In our setting, the state representation additionally includes this temporal component $Y$ that records the unselected duration for each variable—that is, the number of time steps since a variable was last selected or observed, thereby improving decision-making when integrated with the \textbf{causal statistic}. To update the state component $Y_{j,t}$, we track how long data stream $j$ has remained unobserved. Specifically, if variable $j$ is selected for observation at time $t$, then $Y_{j,t}$ is reset to $0$; otherwise, as variable $j$ is unselected, $Y_{j,t}$ is incremented by one from the previous time step. Let $\mathcal{I}(t)$ denote the set of data stream indices selected for observation at time $t$. The state component $Y_{j,t}$ is then updated as:
$$
Y_{j, t} = 
\begin{cases}
0 & \text{if } j \in \mathcal{I}(t), \\
Y_{j, t-1} + 1 & \text{otherwise}.
\end{cases}
$$

The inclusion of this temporal selection history enhances the state informativeness without violating the Markov property.

\paragraph{Causal State} The state space $\mathcal{S}$ is consisted of the above 3 pieces of information, and it can be viewed as a $3 \times p$ matrix as $\mathcal{S}(t)=\left(\begin{array}{llll}\Lambda_{1, t} & \Lambda_{2, t} & \ldots & \Lambda_{p, t} \\ \phi_{1, t} & \phi_{2, t} & \ldots & \phi_{p, t} \\ Y_{1, t} & Y_{2, t} & \ldots & Y_{p, t}\end{array}\right)$. This state representation assists the network in selecting data streams based on following criteria:  
\textbf{(1)} A larger local statistic indicates a higher likelihood of mean shift, which tends to result in a higher $Q$-value;  
\textbf{(2)} Similarly, a higher causal statistic implies that the variable has a more significant causal propagation effect on other data streams, which also contributes to a higher $Q$-value. Higher $Q$-value is related to the action implemented by the network then. \textbf{(3)} By tracking how long a variable has been unobserved ($Y_{j,t}$), the Causal DQ can prioritize exploration of neglected variables, especially those with high \textbf{causal effect}, thus improving both detection performance and sample efficiency.


\paragraph{Action} We represent the action space as a binary vector $\mathcal{A}_t=\left(a_{1, t}, a_{2, t}, \ldots, a_{p, t}\right)$, where each $a_{i, t} \in\{0,1\}$ is a binary indicator variable representing whether the $i$-th data stream is selected for observation at time $t$. The action space is constrained such that exactly $m$ data streams are observed at each step, i.e., $\sum_{j=1}^p a_{j, t}=m$. The $Q$-network outputs $Q$-values for each individual data stream, then selects the top-$m$ data streams with the highest $Q$-values after training. This approach avoids explicitly enumerating all $\binom{p}{m}$ possible combinations, reducing computational complexity from exponential to linear in $p$.


\subsection{Causal Reward}\label{tab:reward}

To enhance the explainability of reinforcement learning, we consider the causal effect from action to reward. In this work, we incorporate the causal information of data stream selection (i.e., actions) and states into the reward function $R$.

Typically, during periods when mean shifts occur in the data streams, a reward is assigned to the $Q$-network once it performs a preferred action \citep{li2025online} (in our case, selecting any single data stream with a mean shift). 
In anomaly detection for streamed monitoring, we want to select as many mean shifted data streams as possible; if we award the same reward regardless of how many anomalies are found, it dilutes the network’s motivation to focus on detecting them. Therefore, the more shifted streams an action selects, the stronger its positive causal effect on the reward. Thus, we introduce the causal mask for representing the action to reward, denoted as $M^{a \rightarrow r}$, where $M \in \mathbb{R}^{p\times 1}$. Likewise, the state $s$ has an indirect causal effect on reward assignment, since it is updated based on the chosen action at each time step. The causal mask of the state is denoted as $M^{s \rightarrow r}$. Each element in this mask is binary, i.e.,\ $M_i \in \{0,1\}$ for $i \in \{1,\ldots,p\}$, where $M^{a \rightarrow r }_i = 1$ indicates that action $a_i$ selects a stream with a mean shift and thus has a causal effect on the reward. Similarly, this action $a_i$ is associated with the state component $s_i$ in $M_i^{s \rightarrow r} = 1$. 

Concretely, during a mean shift in the data streams starting at time step $T$, we define the instantaneous reward $r_t$ at each time $t \in [T, T + e]$, where $e$ denotes the duration of the anomaly (i.e., the expected length of time the anomaly persists). The reward $r_t$ is defined as:

$$r_t= \begin{cases}g(M^{s \rightarrow r} \odot s_t, M^{a \rightarrow r } \odot a_t) & \text { if mean shifts are correctly detected } \\  &  \\ U & \text { if the agent fails to detect any true mean shift}\end{cases}$$ where $\odot$ denotes the element-wise product. In this work, we specify the function $g$ as $g = \sum_{i=1}^p a_{t,i} \cdot y_i + w_i \cdot s_{t,i},$
where $y_i, w_i \in \mathbb{R}^+$ are predefined reward values: they are positive if the $i$-th stream has experienced a true mean shift, and $0$ otherwise. The term $s_{t,i} \in \{0,1\}$ is a state indicator associated with the action $a_i$, indicating whether the $i$-th data stream is selected at time $t$. Similarly, $a_{t,i}$ represents the action of selecting the $i$-th data stream at time $t$, and if that stream has undergone a mean shift, the action has a causal effect on the resulting reward.  Importantly, we assign a large negative constant $U$ (e.g., $-20$) to $r_t$ if the network fails to detect any mean shifted variables, thereby penalizing completely incorrect decisions. Overall, supposed the mean shift occurs at time $T$, the assigned reward at time epoch $t$ is summarized as: 

$$R_t= \begin{cases}r_1 & \text { if } t<T \\ r_t & \text { if } T \leq t \leq T+e \\ r_2 & \text { if } t>T+e\end{cases}$$ where $r_1$ and $r_2$ are baseline constant rewards assigned to prior and post mean shift periods, respectively.

\subsection{Offline Alarm Triggering}\label{tab:Offline}
The Causal DQ is capable of selecting $m$ variables out of $p$ in order to maximize the detection of data streams that exhibit a mean shift. As mentioned in Section~\ref{sec:causalSA}, we compute the aggregated local statistic $ \Lambda(n)=\sum_{i=1}^m \hat{\Lambda}_i(n)$ at each time step. Once $\Lambda(n)$ exceeds its corresponding $\chi^2_p$ threshold, an alarm is triggered. For example, when $p=10$ and $m=5$, at the 95\% confidence level we have $\chi^2_{10}=18.3$; thus, the alarm is triggered when $\Lambda(n)>18.3$. During testing, the $m$ data streams are directly determined by selecting the ones corresponding to the top $m$ ranked $Q$-values, since the network has learned to identify mean-shifted data streams based on the accumulated rewards during training.

\section{Theoretical Results}\label{sec:maththeory}

In this section, we analyze the theoretical properties of our proposed Causal DQ framework for anomaly detection. Specifically, we derive rigorous error bounds and establish the convergence rate under some mild assumptions. Due to space constraints, detailed proofs are provided in the Appendix.

We first establish a lemma demonstrating that the proposed Causal DQ framework still operates within the framework of Markov Decision Processes (MDPs). Consequently, the introduced \textbf{causal entropy} regularizer $\mathcal H_c(\pi(\cdot\mid s'))$ \textbf{preserves the contraction property} of the Bellman operator. This preservation guarantees the \textbf{convergence} of Causal DQ under the standard MDP assumptions.

\newtheorem{lemma}{Lemma}[section]
\begin{lemma}\label{lem:lemma1}
Consider the \textbf{causal entropy–regularized} Bellman operator defined by
$$
\mathcal T_c^\pi Q(s,a)
= r(s,a)
+ \gamma\,\mathbb E_{s'}\Bigl[\,
   \sum_{a'}\pi(a'\mid s')\,Q(s',a')
   \;-\;\tau\,\mathcal H_c\bigl(\pi(\cdot\mid s')\bigr)
\Bigr].
$$
Then for any two bounded action‐value functions $Q$ and $Q'$,
$$
\bigl\|\mathcal T_c^\pi Q \;-\;\mathcal T_c^\pi Q'\bigr\|_\infty
\;\le\;
\gamma\,\|\,Q - Q'\|_\infty,
$$
i.e.\ $\mathcal T_c^\pi$ is a $\gamma$–contraction under the sup‐norm.
\end{lemma}

The \textbf{Causal DQ} algorithm employs a parameterized $Q$-network rather than tabular $Q$-learning. We apply stochastic gradient descent (SGD)  to optimize the objective function (Eq.\ref{eq:entorpy-TD}) in Section \ref{tab:causalEntropy}: $$\mathcal{L}(\theta) = \mathbb{E}_{(s, a, r, s') \sim \text{replay}} \left[ (y - Q(s, a; \theta) )^2 \right] - \alpha \cdot \mathbb{E}_s[\mathcal{H}_c\bigl(\pi(\theta\mid s)\bigr)].$$

We make the following assumption regarding the boundedness of the stochastic gradient of this loss function.  

\newtheorem{assumption}{Assumption}[section]

\begin{assumption}[Bounded Stochastic Gradients]
Let $Q_{\theta}$ be a neural Q-network. The stochastic gradient $\nabla_\theta \mathcal{L}(\theta)$ computed on a batch $\mathcal{B}$ satisfies:  
$$\mathbb{E}_{\mathcal{B}, \xi} \left[ \left\| \nabla_\theta \mathcal{L}(\theta; \mathcal{B}, \xi) \right\|^2 \right] \leq G, \quad \forall \theta$$ where $\mathcal{B} = (s_t, a_t, r_t, s_t') $ is sampled batch from replay buffer at each time step $t$, and $\xi$ encompasses randomness in exploration–exploitation, reward, and size of batch sampling. The constant $G > 0$ is independent of $\theta$, $\mathcal{B}$, and $\xi$. 
\end{assumption}

This loss function $\mathcal{L}(\theta_{\text{online}})$ is regularized by the causal entropy $\mathcal{H}_c$. To ensure optimization stability and convergence, we aim to verify that the entropy regularizer $\mathcal{H}_c$ is a convex function of the policy parameters.
 
 \newtheorem{proposition}{Proposition}[section]
\begin{proposition} \label{prop:th1.2}
The negative causal entropy $-\mathcal{H}_c(\pi(\cdot \mid s))$ is convex, and $\mathcal{H}_c(\pi(\cdot \mid s))$ is concave. 
\end{proposition}

The above theoretical results demonstrate that our proposed \textbf{Causal DQ} preserves the fundamental properties of RL under the MDP framework, including the Markov property and the contraction of the Bellman operator, thereby ensuring convergence.

By explicitly incorporating causal-informed pattern into each stage of learning and introducing the causal entropy $\mathcal{H}_c$ as a regularization term in the $Q$-value update, the resulting optimal $Q$-function $Q^*$ differs from that of the traditional (non-causal) formulation. Therefore, we provide new boundary for the $Q^*$ of \textbf{Causal DQ}. 

\newtheorem{theorem}{Theorem}[section] 

\begin{theorem} \label{thm:th1.1}
Consider a \textbf{causal entropy-regularized} $\operatorname{MDP}\left\langle\mathcal{S},  \mathcal{H}_c, \mathcal{A}, r, \pi\right\rangle$ with optimal value function $Q^*(s, a)$.  As  $\mathcal{H}_c\bigl(\pi(\cdot\mid s)\bigr)=-\mathbb{E}_{a_i\in \mathcal{A}}\left[\sum_{a_i}\mathbf{C}_{a_i\rightarrow r\mid s}\,\pi\bigl(a_i\mid s\bigr)\log \pi\bigl(a_i\mid s\bigr)\right]$, $Q^*(s, a)$ is bounded by:

$\begin{aligned} & Q^*(s, a) \geq r(s, a)+\gamma \mathrm{E}\left[\max _{a^{\prime} \in  \mathcal{A}\left(s^{\prime}\right)} Q\left(s^{\prime}, a^{\prime}\right)\right], \\ & Q^*(s, a) \leq r(s, a)+\gamma\mathrm{E}\left[\max _{a^{\prime} \in \mathcal{A}\left(s^{\prime}\right)} Q\left(s^{\prime}, a^{\prime}\right)\right]+\frac{\gamma}{1-\gamma} \frac{\log \sum_{a_i} \mathbf{C}_{a_i \rightarrow r \mid s}}{\tau}. \end{aligned}$

\end{theorem}

\newtheorem{remark}{Remark}[section]
\begin{remark}\label{remark 1}
This theorem (Theorem \ref{thm:th1.1})  provides a \textbf{tighter upper bound} on the optimal value function $Q^*(s, a)$ by explicitly leveraging causal information. In contrast to the non-causal upper bound-denoted as $Q_{\mathrm{nc}}^*$ which depends on the cardinality of the action space $|\mathcal{A}|$.
\end{remark}

In this paper, we consider the data streams with anomaly when computing the causal reward (see Section~\ref{tab:reward}). As a result, the magnitude of the causal mask—that is, the number of actions with positive causal effect on reward—is strictly smaller than the total action space size, since the number of partially placed sensors is always less than the total number of variables. 

\begin{remark}\label{remark 2} The magnitude of  causal mask in reward computing satisfies $ \sum_{a_i} \mathbf{C}_{a_i \rightarrow r \mid s}\leq m <|\mathcal{A}|=\binom{p}{m}$ as $1 \leq m<p$
where $m$ is total number of monitored streams and $\binom{p}{m}$ is the cardinality of the action space $|\mathcal{A}|$. This implies that the error bounds for the optimal value function $Q^*$ in Causal DQ are \textbf{tighter than} those for $Q_{nc}^*$ in non-causal $Q$-network.
\end{remark}

Furthermore, the error bound between the learned $Q$-value at time step $t$, denoted as $Q_t$, and the optimal value $Q^*$, becomes tighter after incorporating causality into the learning process.
\newtheorem{corollary}{Corollary}[section]

\begin{corollary}\label{tab:Corollary 1}

Consider a \textbf{causal entropy-regularized} MDP
$\langle \mathcal{S}, \mathcal{H}_c, \mathcal{A}, r, \pi\rangle$
with optimal value function $Q^*(s,a)$. As shown in Theorem~\ref{thm:th1.1}, $Q^*(s,a)$ admits the stated bound. The error bound associated with convergence of the learned function $Q_t$ toward $Q^*(s,a)$ at each time step $t$ can then be summarized as follows:

$$\left\|Q_t-Q^*\right\|_{\infty} \leq \gamma^t\left\|V_{0}-V^*_c\right\|_{\infty}+\frac{\gamma}{1-\gamma}\frac{\log\sum_{a_i}\mathbf{C}_{a_i \rightarrow r \mid s}}{\tau}$$ where $V_0$ is the initial value when $t=0$ and $V^*_c = \frac{1}{\tau w(s)}\left[\mathcal{H}_c(\pi(\cdot \mid s))+\tau \sum_{a_i} \mathbf{C}_{a_i \rightarrow r \mid s} \pi(a_i \mid s) Q^*(s, a_i)\right]$ as $w(s)=\sum_{a_i} \mathbf{C}_{a_i \rightarrow r \mid s} \pi(a \mid s)$.

\end{corollary}

\begin{remark}
The error bound in Corollary~\ref {tab:Corollary 1} reveals that it depends on the sum of causal mask $\sum_{a_i} \mathbf{C}_{a_i \rightarrow r \mid s}$ rather than the size of the entire action space $|\mathcal{A}|$. This substitution implies a \textbf{tighter error bound} compared to that of the non-causal $Q$-network. 
\end{remark}

As $t \rightarrow \infty$, the error bound converges to a more specific value, as characterized in Corollary \ref{coro:corallary2}. This limit further highlights the benefit of incorporating causal information, since the bound depends on the causal mask rather than the full action space size.

\begin{corollary}\label{coro:corallary2}

From the Corollary~\ref{tab:Corollary 1} and \cite{pan2019reinforcement}, we can further have a more specific error bounds when $t \rightarrow \infty$ as: 

$$\lim _{t \rightarrow \infty}\left\|Q_t-Q^*\right\|_{\infty} \leq \min \left\{\frac{\gamma\log \sum_{a_i} \mathbf{C}_{a_i \rightarrow r \mid s}}{\tau(1-\gamma)}, \frac{2 g(\cdot)}{(1-\gamma)^3}\right\}$$ where $g(\cdot)$ is the \textbf{causal reward} function defined in Section~\ref{tab:reward}, which takes as input the predefined reward and the causal mask capturing the causal relationship between action $a$ and state $s$ to reward.
\end{corollary}

\begin{remark}
When $t \rightarrow \infty$, the error bound $\left\|Q_t - Q^*\right\|_{\infty}$ does not converge to zero. This indicates that the Causal DQ introduces a bias due to the causal entropy regularization.
\end{remark}

\begin{remark}
Corollary~\ref{coro:corallary2} shows that adding an entropy regularization term to the Bellman operator can promote exploration but also introduces bias. Notably, this bias is influenced by the causal mask $\sum_{a_i} \mathbf{C}_{a_i \rightarrow r \mid s}$ rather than the size of the action space $|\mathcal{A}|$. While the causal entropy term introduces asymptotic bias, it \textbf{reduces variance during early training} and enables the $Q$-network to \textbf{perform more effective exploration under causal constraints}. Comprehensive exploration over causality is particularly important to prevent the algorithm from converging to sub-optimal solutions, especially in anomaly detection scenarios with partially observed sensor placements. In such settings, sub-optimal sensor configurations may cause anomalies to remain undetected for extended periods.
\end{remark}

\begin{remark}
Corollary \ref{coro:corallary2} further characterizes the asymptotic error bound as $t \rightarrow \infty$, demonstrating that incorporating causality effectively \textbf{reduces the long-term approximation error} throughout the entire process of updating the $Q$-network.
\end{remark}

A tighter error bound between $Q_t$ and $Q^*$ implies that, in Causal DQ, the learned $Q$-values require \textbf{fewer time steps to converge} to the optimal $Q^*$ compared to the non-causal $Q$-network update. Therefore, we derive the number of convergence time steps $T$ required for $Q_t$ to approach the optimal $Q^*$, to provide a more explicit characterization of the convergence behavior.

\begin{lemma}\label{lem:lemma2}
From Corollary \ref{tab:Corollary 1} and \ref{coro:corallary2}, we can conclude that the convergence time steps $T$ required for $Q_t$ to approach $Q^*$ within a precision $\varepsilon$—where the residual error is solely due to the unavoidable bias introduced by the causal mask $\sum_{a_i} \mathbf{C}_{a_i \rightarrow r \mid s}$ (i.e., $\left\|Q_t-Q^*\right\|_{\infty} \leq \varepsilon$ and $\varepsilon-\frac{\gamma}{1-\gamma} \cdot \frac{\log \sum a_i \mathbf{C}_{a_i \rightarrow r \mid s}}{\tau} \rightarrow 0^{+}$)—is asymptotically bounded by:

$$T = \mathcal{O}\left( \log\left( \frac{1}{\varepsilon - \frac{\gamma}{1 - \gamma} \cdot \frac{\log \sum_{a_i} \mathbf{C}_{a_i \rightarrow r \mid s}}{\tau}} \right) \right).$$

\end{lemma}

\begin{remark}
The convergence time steps in \textbf{Causal DQ} depend on the sum of causal mask $\sum_{a_i} \mathbf{C}_{a_i \rightarrow r \mid s}$, which is \textbf{smaller than} that of the non-causal case as $ \sum_{a_i} \mathbf{C}_{a_i \rightarrow r \mid s} < |\mathcal{A}|$. 
\end{remark}

\begin{remark}
Lemma~\ref{lem:lemma2} indicates that the proposed \textbf{Causal DQ} algorithm achieves \textbf{exponential convergence} toward a biased term induced by causal entropy. This convergence is \textbf{faster than} that of the corresponding bias term induced by \textbf{non-causal} action-space entropy (i.e., related to $|\mathcal{A}|$).
\end{remark}

\begin{remark}
When relaxing the precision requirement from accounting only for the bias induced by the causal mask to allowing additional error, the convergence time $T$ is asymptotically bounded by:

$$T=\mathcal{O}\left(\log \left(\frac{\left\|V_0-V_c^*\right\|_{\infty}}{\varepsilon-\frac{\gamma}{1-\gamma} \cdot \frac{\log \sum_{a_i} \mathbf{C}_{a_i \rightarrow r \mid s}}{\tau}}\right)\right)$$

Therefore, according to Lemma~\ref{lem:lemma2}, achieving the minimum possible error $\frac{\gamma\log \sum_{a_i} \mathbf{C}_{a_i \rightarrow r \mid s}}{\tau(1-\gamma)}$,  which is solely determined by the causal mask, would require the number of iterations  $T \rightarrow \infty$ as $\varepsilon- \frac{\gamma\log \sum_{a_i} \mathbf{C}_{a_i \rightarrow r \mid s}}{(1-\gamma)\tau} \rightarrow 0$.
However, this is impractical in real-world applications. Therefore, by allowing a slightly higher total error, we can reduce the required convergence time  $T$, which in turn lowers computational cost.  

\end{remark}

\begin{remark}
The convergence precision deteriorates when the total causal influence $\sum_{a_i} \mathbf{C}_{a_i \rightarrow r \mid s}$ increases or when the temperature parameter $\tau$ decreases, since both factors amplify the bias. Additionally, a higher discount factor $\gamma$ not only slows down the convergence rate but also raises the bias floor, which suggests that $\gamma$ should be chosen within a reasonable range to balance convergence speed and solution accuracy.
\end{remark}

Finally, we provide a finite-time error bound for our proposed \textbf{causal entropy} $Q$-value, expressed as a function of the time step $t$ and other system parameters. Our analysis adopts techniques from control theory, particularly those based on non-linear switching systems \citep{lee2023discrete, jeong2024finite}. 

We follow the scaled unit reward assumption proposed in \cite{jeong2024finite}, which assumes that the reward is uniformly bounded by 1.
\begin{assumption}\label{assum:reward}
The magnitude of the reward signal is uniformly bounded by a constant $R_{\max } \leq 1$ such that:
$\left|r\left(s, a, s^{\prime}\right)\right| \leq R_{\max }$, $\forall\left(s, a, s^{\prime}\right) \in \mathcal{S} \times \mathcal{A} \times \mathcal{S}$.
\end{assumption}

\begin{theorem} \label{thm:ExpectError bounds}
The finite time error bound for our proposed \textbf{causal entropy} $Q$-value is: 

$$\begin{aligned}
\mathbb{E}\left[\left\|Q_t-Q^*\right\|_{\infty}\right]
\leq 
& \frac{4 \alpha \gamma \omega_{\max }|\mathcal{A}|}{1-\gamma} \cdot t \cdot \rho^{t-1}+\frac{2 \sqrt{6} \alpha^{1 / 2} \gamma \omega_{\max }|\mathcal{A}|^{1 / 2}}{\omega_{\min }^{3 / 2}(1-\gamma)^{5 / 2}}\\
& +\frac{ \log \sum_{a_i} \mathbf{C}_{a_i \rightarrow r \mid s}}{\tau}\left[\frac{2 \gamma^2 \omega_{\max }^2}{\omega_{\min }^2(1-\gamma)^2}+\frac{1}{\omega_{\min }(1-\gamma)}+\alpha \gamma \omega_{\max }|\mathcal{A}|^{1 / 2} \sum_{i=0}^{t-1} \rho^{t-i-1}\right] \\
& +\left(|\mathcal{A}|^{2 / 3} \cdot\left(\frac{2}{1-\gamma}\right)^2 \cdot \rho^{2 t}+\frac{6 \alpha|\mathcal{A}|}{\omega_{\min }(1-\gamma)^3}\right)^{1 / 2}
\end{aligned}$$

where $\omega \in (0,1)$ represents the empirical sampling weight of the transition probability associated with a particular state-action pair. Its minimum and maximum values are denoted by $\omega_{\min}$ and $\omega_{\max}$, respectively. We define $\rho := 1 - \alpha \, \omega_{\min} (1 - \gamma)$.
\end{theorem}

We can separate this error bound into four terms as follows:

$\begin{aligned} \mathbb{E}\left[\left\|Q_t-Q^*\right\|_{\infty}\right] \leq & \underbrace{\frac{4 \alpha \gamma \omega_{\max }|\mathcal{A}|}{1-\gamma} \cdot t \cdot \rho^{t-1}}_{\text {(I) }}+\underbrace{\frac{2 \sqrt{6} \alpha^{1 / 2} \gamma \omega_{\max }|\mathcal{A}|^{1 / 2}}{\omega_{\min }^{3 / 2}(1-\gamma)^{5 / 2}}}_{\text {(II) }} \\ & +\underbrace{\frac{\log \sum_{a_i} \mathbf{C}_{a_i \rightarrow r \mid s}}{\tau}\left[\frac{2 \gamma^2 \omega_{\max }^2}{\omega_{\min }^2(1-\gamma)^2}+\frac{1}{\omega_{\min }(1-\gamma)}+\alpha \gamma \omega_{\max }|\mathcal{A}|^{1 / 2} \sum_{i=0}^{t-1} \rho^{t-i-1}\right]}_{\text {(III) }} \\ 
& + \underbrace{\left(|\mathcal{A}|^{2 / 3} \cdot\left(\frac{2}{1-\gamma}\right)^2 \cdot \rho^{2 t}+\frac{6 \alpha|\mathcal{A}|}{\omega_{\min }(1-\gamma)^3}\right)^{1 / 2}.}_{\text {(IV) }}
\end{aligned}$

\begin{remark}
When $t \rightarrow \infty$, term \text{(I)} vanishes (i.e., goes to $0$), while terms \text{(III)} and \text{(IV)} converge to constants. These constants are determined by the size of the action space $|\mathcal{A}|$, the causal mask term $\log \sum_{a_i} \mathbf{C}_{a_i \rightarrow r \mid s}$, the transition probability $\omega$ (which is influenced by the exploration sampling policy), and parameters $\gamma$ and $\tau$. 
\end{remark}

This finite-time error bound (Theorem~\ref{thm:ExpectError bounds}) also reveals that the bias introduced by the causal entropy does not vanish as $t \rightarrow \infty$. Furthermore, this bias term is directly linked to the  the exploration–exploitation trade-off and achievable accuracy. Specifically, while a larger value of $\omega$ can reduce the bias term and improve the precision of convergence, it also indicates that the $Q$-network predominantly samples certain state-action pairs (high sampling probability) while exhibiting low stochasticity in exploring others. This limited exploration introduced by trade-off often results in sub-optimal solutions in reinforcement learning. Moreover, a larger action space $|\mathcal{A}|$ and stronger cumulative causal influence $\sum_{a_i} \mathbf{C}_{a_i \rightarrow r \mid s}$ can enlarge the error term. This reflects an unavoidable trade-off introduced by the causal entropy regularization. 

Notably, this finite-time error bound is smaller in our proposed Causal DQ compared to the non-causal version, as reflected in term \text{(III)}.

\begin{remark}
Term \text{(III)} in the \textbf{non-causal} is  driven by the action space $|\mathcal{A}|$. This makes it larger than the corresponding term in the causal version, which is instead driven by the causal mask $\log \sum_{a_i}\mathbf{C}_{a_i \rightarrow r \mid s}$.
\end{remark}

Consequently, \textbf{both the error bound and convergence time of Causal DQ are strictly smaller than} those of the non-causal case. This improved convergence rate advantage is also demonstrated in the experimental results and real-world applications (see Section~\ref{sec:Experiment} and Section~\ref{sec:realcase}).

\section{Experimental Studies}\label{sec:Experiment}


















We simulate various anomaly scenarios to illustrate the performance of our causality‐informed anomaly detection algorithm. Meanwhile, we compare our approach with the following state-of-the-art baselines:

\begin{enumerate}

    \item \textbf{Non-Causal}: First, we remove all causal components from our proposed Causal DQ—leaving only the $Q$-network—to evaluate its performance in the non-causal setting.

    \item \textbf{TSSRP:} \citet{zhang2023bandit} proposed a multi‐armed bandit framework for optimal sensor selection that incorporates Bayesian priors which account for potential anomalies in the data streams.

    \item \textbf{ASMP}: \cite{yao2023adaptive} proposed an adaptive sampling framework (ASMP) for monitoring multi‐profile datasets under partially observable sensors. This framework employs multivariate functional principal component analysis (MFPCA) to capture the correlations among variables.

    \item \textbf{SDQ}: \cite{li2025online}  employ a double-Q network to select variables that exhibit mean shifts, adopting a reward function and an action-selection strategy distinct from our proposed approach.

\end{enumerate}

\paragraph{Simulated Datasets.} We simulate multiple datasets of $p$-dimensional observations over $t=200$ time steps. Since the variables were generated independently, no causal relationships exist at prior. To impose a ground truth causal structure, we sample a directed acyclic graph (DAG) via an Erdős-Rényi (ER) model \citep{kocaoglu2020applications, lopez2022large, yang2023reinforcement}. Specifically, over all variables $X = \{X_1, \dots, X_p\}$ and each ordered pair of nodes $(i, j) \in \{1, \dots, p \}$, we first assign a random topological ordering $\rho$ of the nodes, where $\rho(i)$ denotes the position of node $X_i$ in this ordering (i.e., an integer from $1$ to $p$). This ensures acyclicity by allowing a directed edge $X_i \rightarrow X_j$ only if $\rho(i)<\rho(j)$, meaning a variable can only influence those that come later in the ordering.

Next, for each ordered pair \((i,j)\) with $\rho(i)<\rho(j)$, we include the directed edge $X_i\to X_j$ independently with probability $\alpha \leq 1$, yielding a DAG with random assigned and directed acyclic edges.

Finally, We produce an acyclic graph with the random expected number of edges as the above construction. 



\begin{figure}[H]
  \centering
  
  \begin{subfigure}[b]{0.9\textwidth}
    \centering
    \includegraphics[width=\textwidth]{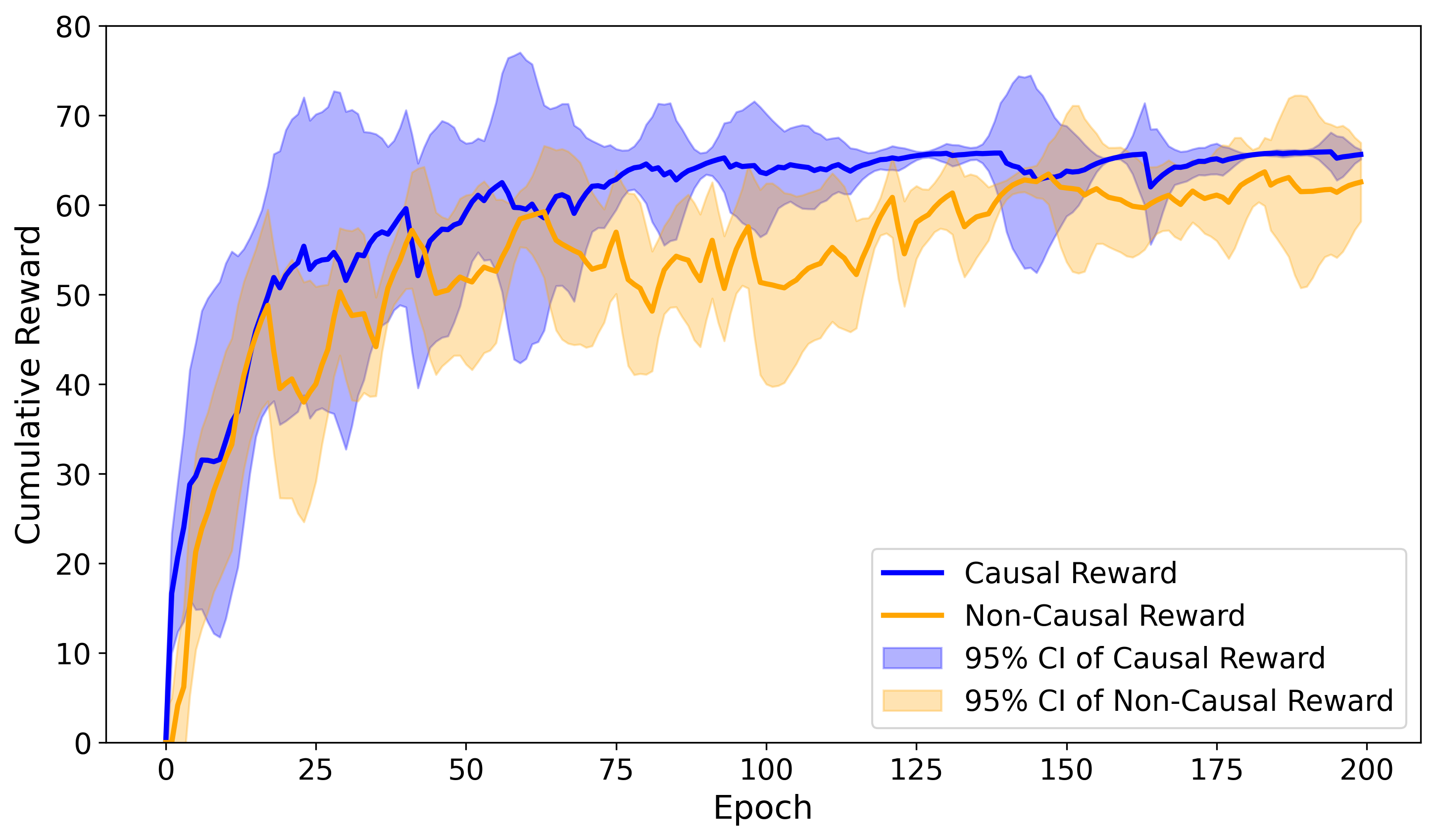}
    \caption{$p = 10$}
    \label{fig:p10}
  \end{subfigure}
  \hfill
  
  \begin{subfigure}[b]{0.9\textwidth}
    \centering
    \includegraphics[width=\textwidth]{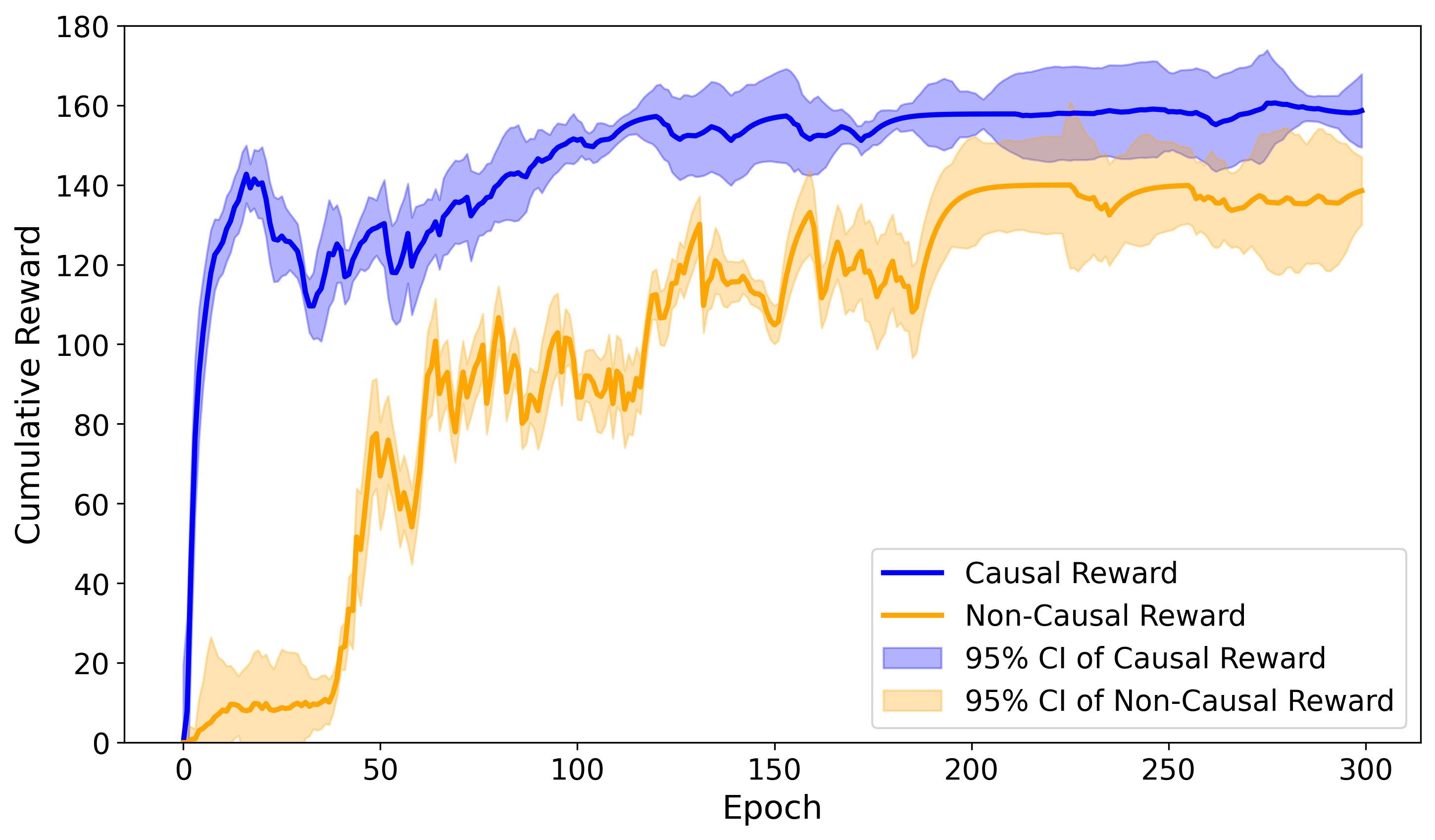}
    \caption{$p = 50$}
    \label{fig:p50}
  \end{subfigure}
  \hfill
  
  \begin{subfigure}[b]{0.9\textwidth}
    \centering
    \includegraphics[width=\textwidth]{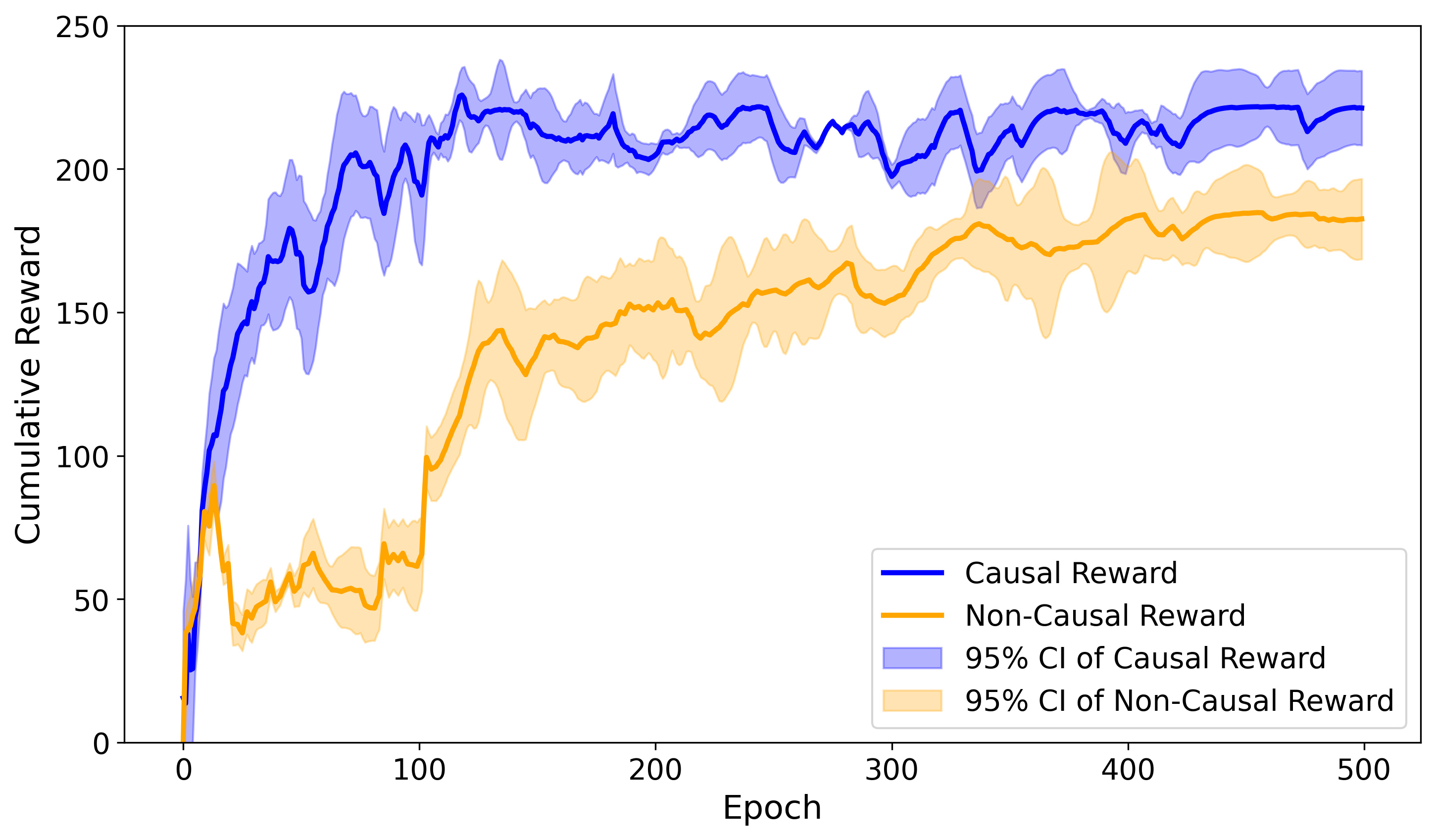}
    \caption{$p = 100$}
    \label{fig:p100}
  \end{subfigure}

  \caption{Total Reward Comparison between Causal and Non-Causal DQ with different values of $p$.}
  \label{fig:three_DQ_r}
\end{figure}

\paragraph{Training Network.} We train the Causal DQN separately on each anomaly detection scenario ($p$=10, 50, and 100). Each network comprises three hidden layers, and each layer contains 256 neurons.  For our Causal DQ, we recommend performing between 200 and 400 iterations of Q-function updates per episode to ensure convergence. 
In parallel, we train Non-Causal DQ networks under the same scenarios. Figure \ref{fig:three_DQ_r} compares the cumulative rewards achieved by both approaches after training. 

As shown in Figure \ref{fig:three_DQ_r}, our proposed Causal DQ converges at a rate similar to Non-Causal DQ when $p=10$, $m =6$. However, as the dimension of data streams increases to $p=50$, $m =12$ and $p=100$, $m=22$, Causal DQ converges faster and more stable than the standard $Q$-network without causality. For instance, at $p=100$, the Causal DQ's cumulative reward reaches its maximum around epochs 100-200, whereas the Non-Causal $Q$-network only converges by around epoch 400.

The results shown in Figure~\ref{fig:three_DQ_r} are consistent with our \textbf{theoretical analysis} (Section~\ref{sec:maththeory}), as convergence occurs at a faster exponential rate  and the error bounds exhibit an exponential decay trend. The convergence rate of Causal DQ becomes more obviously when the dimensionality increases,  especially for $p=50$ and $p=100$, when the dimensional is increased. Moreover, since Causal DQ has smaller error bounds between the estimated $Q_t$ at each time step $t$ to real optimal value $Q^*$, it can achieve higher cumulative rewards. This improvement comes from its ability to identify more mean-shift points by leveraging causal information.

\subsection{Noise-free Mean Shift}\label{sec:Noise-free}

We first examine two straightforward settings in this section. We set the first $k$ variables with mean shifts $\mu_{c}$ as: (a).  $\mu_{c}=[\underbrace{1, \ldots, 1}_k, 0, \ldots, 0] \times \delta$ and (b). $\mu_c=[\underbrace{1,-1,1, \ldots, 1,-1}_k, 0, \ldots, 0] \times \delta$, where $\delta$ represents the shifted values. We examine $k = 5$ for $p =10$, $k = 10$ for $p=50$, and $k = 20$ for $q=100$. The number of sensors is then chosen as $ m= 6$ for $p =10$, $m = 12$ for $p=50$, and $m = 22$ for $q=100$, as the number of samples is fixed to $t = 200$ time steps. We report the results of each scenario in Table~\ref{tab:p=10non-noise}, Table~\ref{tab:p=50non-noise}, and Table~\ref{tab:p=100non-noise} with their standard errors in parentheses. The resulted ADDs are derived from 100 simulation replications. 

As shown in the reported results, our Causal DQ consistently achieves the shortest ADD across all dimensionalities.  While all baseline methods perform well under large mean shifts (e.g., $\delta = 2$), their detection power drops significantly when the shift becomes subtle (e.g., $\delta = 0.25$). Notably, our Causal DQ retains superior detection power even under the weak shifts  (e.g., $\delta = 0.25$ or $\delta = 0.5$) because of its ability to leverage causality. By focusing on causally-relevant variables rather than all raw dimensions, Causal DQ can more effectively detect even slight mean shift.

\begin{table}[H]
\centering

\resizebox{\textwidth}{!}{%
  \begin{tabular}{@{} c r r r r r @{\quad} c r r r r r @{}}
  \toprule
   & \multicolumn{5}{c}{Case (a)} & & \multicolumn{5}{c}{Case (b)} \\
  \cmidrule(lr){2-6} \cmidrule(lr){8-12}
  $\delta$ 
   & Causal DQ & Non-Causal & TSSRP & ASMP  & SDQ 
   & & Causal DQ & Non-Causal  & TSSRP & ASMP & SDQ \\
  \midrule
  0.25 & \textbf{62.4} (5.2) & 71.3 (4.7) & 71.6 (7.5) & 65.4 (6.8) & 83.9 (4.6)
       & & \textbf{68.5} (3.2) & 74.2 (6.1) & 74.6 (5.4) & 62.1 (3.7) & 79.9 (5.4) \\
  0.50 & \textbf{19.5} (3.2) & 27.4 (2.3) & 38.9 (1.8) & 30.8 (0.3)  & 28.6 (3.1)
       & & \textbf{16.3}(3.6) & 28.1 (2.7)  & 30.4 (1.5) & 29.9 (2.8) & 24.6 (3.4) \\
  1.00 &  \textbf{10.8} (2.1) & 13.3 (2.4) &  12.6 (2.3) & 11.9 (1.7) & 12.2 (2.3)
       & & \textbf{8.3} (0.5) & 11.3 (0.8) &  10.5 (1.2) & 10.9 (1.4) & 11.2 (0.7) \\
  1.50 &  \textbf{4.9} (0.7) & 6.3 (0.1) &  9.4 (0.6) & 8.0 (2.0) & 7.7 (1.8)
       & &  \textbf{5.0} (0.1) & 8.6 (0.7) &  10.7 (0.8) & 9.8 (1.3) & 9.1 (1.2) \\
  2.00 &  \textbf{4.2} (0.2) & 6.2 (0.2) &  7.8 (0.4) & 7.4 (0.7) & 6.9 (0.6)
       & &  \textbf{3.2} (0.2) & 5.7 (0.3) &  5.8 (0.5) & 6.4 (0.8) & 7.1 (0.5) \\
  \bottomrule
  \end{tabular}%
} 
\caption{Average Detection Delays (ADDs) of different methods for change pattern ($p=10$).}
\label{tab:p=10non-noise}
\end{table}

\begin{table}[H]
\centering

\resizebox{\textwidth}{!}{%
  \begin{tabular}{@{} c r r r r r @{\quad} c r r r r r @{}}
  \toprule
   & \multicolumn{5}{c}{Case (a)} & & \multicolumn{5}{c}{Case (b)} \\
  \cmidrule(lr){2-6} \cmidrule(lr){8-12}
  $\delta$ 
   & Causal DQ & Non-Causal & TSSRP & ASMP  & SDQ 
   & & Causal DQ & Non-Causal  & TSSRP & ASMP & SDQ \\
  \midrule
  0.25 & \textbf{110.5} (7.2) & 115.4 (8.5) & 125.6 (9.2) & 121.4 (8.7) & 120.9 (9.0)
       & & \textbf{104.2} (3.8) & 113.7 (8.3) & 120.6 (9.2) & 112.4 (8.0) & 117.9 (7.7) \\
  0.50 & \textbf{25.7} (1.3) & 29.1 (2.1) & 57.4 (3.3)  & 40.2 (2.3) & 35.6 (2.3)
       & & \textbf{27.3} (8.5) & 31.4 (3.4) & 54.4 (3.0) & 38.1 (4.1) & 40.0 (3.5) \\
  1.00 &  \textbf{12.2} (0.3) & 13.8 (0.2) &  14.5 (0.3)& 13.9 (0.5) & 15.2 (0.6)
       & &  \textbf{11.2} (1.3) & 11.7 (0.4 )&  12.4 (0.5) & 12.9 (0.5) & 14.2 (0.8) \\
  1.50 &  \textbf{5.7} (0.4) & 6.3 (0.6) &  6.5 (0.1) & 6.9 (0.4) & 7.2 (0.2)
       & &  \textbf{6.0}  (0.3) & 6.2 (0.4) &  7.1 (0.2) & 6.8 (0.7)& 6.6 (0.4) \\
  2.00 &  \textbf{5.1} (0.2) & 5.2 (0.1) &  5.8 (0.2) & 5.4 (0.1) & 5.4 (0.1)
       & &  \textbf{5.7} (0.5) & 6.2 (0.1) &  6.7 (0.3) & 5.8 (0.2) & 6.0 (0.4) \\
  \bottomrule
  \end{tabular}%
} 
\caption{Average Detection Delays (ADDs) of different methods for change pattern ($p=50$).}
\label{tab:p=50non-noise}
\end{table}

\begin{table}[H]
\centering

\resizebox{\textwidth}{!}{%
  \begin{tabular}{@{} c r r r r r @{\quad} c r r r r r @{}}
  \toprule
   & \multicolumn{5}{c}{Case (a)} & & \multicolumn{5}{c}{Case (b)} \\
  \cmidrule(lr){2-6} \cmidrule(lr){8-12}
  $\delta$ 
   & Causal DQ & Non-Causal & TSSRP & ASMP  & SDQ 
   & & Causal DQ & Non-Causal  & TSSRP & ASMP & SDQ \\
  \midrule
  0.25 & \textbf{124.3} (8.2) & 127.4 (7.8) & 144.3 (8.7) & 130.8 (6.9)& 133.9 (7.6)
       & & \textbf{127.3} (8.4) & 132.4 (8.9) & 141.5 (9.1) & 145.4 (7.2) & 139.6  (9.0) \\
  0.50 & \textbf{29.2} (3.3) & 33.2 (4.0) & 60.8 (2.0)& 46.7 (3.1) & 40.6 (4.1)
       & & \textbf{28.1} (4.1) & 34.2 (3.9) & 54.7 (5.8) & 46.4 (6.9) & 42.6 (5.2) \\
  1.00 &  \textbf{14.6} (0.3) & 17.2 (0.8)  &  16.3 (0.3) & 15.7 (0.2) &16.8 (0.7)
       & &  \textbf{12.8} (0.4) & 21.3 (0.5)&  14.4 (0.3)  & 13.6 (0.2) &15.9 (0.9) \\
  1.50 &  \textbf{6.4} (0.4) & 7.4 (0.2) &  7.8 (0.1) & 7.0 (0.2) & 6.6 (0.3)
       & &  \textbf{6.8} (0.1) & 7.3 (0.1) &  7.8 (0.1) & 7.6 (0.2) & 8.1 (0.1) \\
  2.00 &  \textbf{5.9} (0.4) & 6.2 (0.2) &  7.6 (0.1) & 6.1 (0.1) & 6.0 (0.2)
       & &  \textbf{6.6} (0.2) & 6.9 (0.3) &  7.0 (0.2) & 7.4 (0.4) & 8.1(0.3) \\
  \bottomrule
  \end{tabular}%
} 
\caption{Average Detection Delays (ADDs) of different methods for change pattern ($p=100$).}
\label{tab:p=100non-noise}
\end{table}

\subsection{Mean Shift with Noise}\label{sec:diffNoise}

To demonstrate the robustness of our Causal DQ approach against noise-induced bias, this section examines data streams subjected to mean-shift perturbations. Because sensor readings may exhibit minor fluctuations due to noise—fluctuations that should not be regarded as anomalies—we base our noisy-scenario design on the noise-free experiments presented in Section~\ref{sec:Noise-free}, e.g. $\mu_{c}=[\underbrace{1, \ldots, 1}_k, 0, \ldots, 0] \times \delta$ + \textit{noise}. We examine $k = 10$ for $p=50$ and $k = 20$ for $q=100$. The number of sensors is then chosen as $m = 12$ for $p=50$ and $m = 22$ for $q=100$. Importantly, when no mean shift is present (e.g. $\delta=0$), random noise should not be mistaken for an anomaly, and no alarm should be triggered. In this section, we fix the number of samples to $t = 200$ time steps.  


We sample the noise from Normal Distribution as: $\mathcal{N}(0,0.05)$, $\mathcal{N}(0,0.1)$ and $\mathcal{N}(0,0.15)$ which is denoted as $\mathcal{\sigma}=0.05$, $\mathcal{\sigma}=0.1$, and $\mathcal{\sigma}=0.15$ in the reported Table~\ref{tab:p=50noise} and Table~\ref{tab:p=100noise}. 

The Causal DQ still presents the best performance (shortest ADD) under all occurrence of mean shift with noise perturbations. Notably, when $\delta = 0$, our Causal DQ does not trigger any alarms as the ADD $= 200$, which demonstrates its robustness to mere noise fluctuations.



\begin{table}[H]
  \centering

  \resizebox{\textwidth}{!}{%
    \begin{tabular}{c|rrrrr|rrrrr|rrrrr}
      \toprule
       & \multicolumn{5}{c|}{$\mathcal{\sigma}=0.05$}
      & \multicolumn{5}{c|}{$\mathcal{\sigma}=0.1$}
      & \multicolumn{5}{c}{$\mathcal{\sigma}=0.15$} \\
      $\delta$
      & Causal DQ & Non-Causal & TSSRP & ASMP  & SDQ 
      & Causal DQ & Non-Causal & TSSRP & ASMP  & SDQ 
      & Causal DQ & Non-Causal & TSSRP & ASMP  & SDQ  \\
      \midrule
      0 & 200 (0) & 200 (0) & 187.4 (4.1) & 188.2 (2.8)
            &  200 (0) & 200 (0) &200 (0) & 194.5 (1.6) & 193.9 (2.8)
            &  200 (0) &  200 (0) &  200 (0) &  197.8 (1.6) &  195.8 (1.6) & 200 (0) \\
      0.25 & \textbf{115.2} (6.0) & 124.2 (5.0) & 170.3 (5.4) & 143.4 (6.3) & 141.3 (5.4)
            &\textbf{118.6} (5.8) & 128.2 (5.4) & 177.5 (5.1) & 155.2 (7.9) & 130.9 (7.5)
            &  \textbf{123.7} (8.9) &  130.2 (9.4) &  186.2 (8.4) &  159.1 (7.0) &  150.3 (8.0) \\
      0.50 & \textbf{27.8} (4.1) & 35.2 (3.2) & 54.6 (4.0) &  48.1 (4.8) & 38.8 (5.7)
            & \textbf{30.2} (2.4) & 41.1 (1.8) & 50.2 (3.4) &  52.8 (4.2) & 42.7 (3.3)
            &  \textbf{31.3} (4.3) &  39.2 (2.1) &  58.2 (5.1) &  56.2 (2.2) &  41.1 (5.1) \\
      1.00 & \textbf{10.7} (0.5) & 12.8 (0.3) &  11.5 (0.3) &  12.9 (0.6) & 12.4 (0.5)
            & \textbf{11.6} (0.8) & 12.3 (0.6) &  12.0 (0.3) &  13.2 (0.7) & 13.4 (0.6)
            &  \textbf{11.7} (0.2) &  12.0 (0.4) &  11.8 (0.6) &  12.1(0.5) &  11.9 (0.5) \\
      2.00 &  \textbf{5.3} (0.3) & 5.7 (0.1) &  5.7 (0.3) &  5.4 (0.1) & 5.5 (0.2)
            &  \textbf{5.6} (0.8) & 6.2 (0.2) &  5.8 (0.3) &  5.7 (0.3) & 6.2 (0.4)
            &   \textbf{5.3} (0.1) &  5.8 (0.2) &  5.7 (0.6) &  6.4 (0.4) &  5.9 (0.1) \\
      \bottomrule
    \end{tabular}}%
      \caption{Average Detection Delays (ADDs) of different methods with \textbf{noise} for change pattern ($p=50$).}
  \label{tab:p=50noise}
\end{table}

\begin{table}[H]
  \centering

  \resizebox{\textwidth}{!}{%
    \begin{tabular}{c|rrrrr|rrrrr|rrrrr}
      \toprule
      & \multicolumn{5}{c|}{$\mathcal{\sigma}=0.05$}
      & \multicolumn{5}{c|}{$\mathcal{\sigma}=0.1$}
      & \multicolumn{5}{c}{$\mathcal{\sigma}=0.15$} \\
      $\delta$
      & Causal DQ & Non-Causal & TSSRP & ASMP  & SDQ 
      & Causal DQ & Non-Causal & TSSRP & ASMP  & SDQ 
      & Causal DQ & Non-Causal & TSSRP & ASMP  & SDQ  \\
      \midrule
      0 & 200 (0.0) & 200 (0.0) & 200 (0.0) & 200 (0.0) & 200 (0.0)
            & 200 (0.0) & 200 (0.0) & 200 (0.0) & 200 (0.0) & 200 (0.0)
            &  200 (0.0) & 200 (0.0) & 200 (0.0) & 200 (0.0) & 200 (0.0)\\
      0.25 & \textbf{121.4} (7.1) & 141.2 (8.2) & 180.1 (9.3) & 152.2 (8.2) & 143.9 (6.7)
            & \textbf{123.5} (5.3) & 137.8 (6.7) & 179.6 (7.9) & 160.4 (6.9) & 153.9 (5.0)
            &  \textbf{125.1} (7.1) &  138.5 (6.1) &  187.2 (6.5) &  166.3 (6.8) & 155.1 (8.6) \\
      0.50 & \textbf{29.1} (2.1) & 38.5 (4.2) & 63.2 (5.3) &  50.3 (5.1) & 42.1 (5.9)
            & \textbf{35.5} (4.0) & 48.2 (5.2) & 69.3 (6.0) &  56.5 (4.4) & 55.9 (5.0)
            &  \textbf{36.4} (2.1) &  49.3 (3.5) &  70.2 (5.2) & 58.1 (5.1)  &  54.8 (4.2) \\
      1.00 & \textbf{12.1} (0.7) & 16.3 (1.2) &  16.5 (1.0) &  15.8 (0.4) & 16.6 (0.2)
            & \textbf{12.2} (0.5) & 15.1 (0.4) &  18.6 (0.1) &  15.1 (0.2) & 18.2 (0.6)
            &  \textbf{13.9} (0.1) &  18.2 (0.4) &  18.9 (0.1) &  16.7 (0.2) &  18.2 (0.2) \\
      2.00 &  \textbf{5.6} (0.3) & 6.0 (0.3) &  7.2 (0.6) &  7.4 (0.2)& 6.3 (0.1)
            & \textbf{5.8} (0.4) & 6.7 (0.1) &  7.6 (0.3) &  8.4 (0.1) & 6.4 (0.1)
            &   \textbf{5.9} (0.1) &  6.3 (0.1) &  7.5 (0.3) &  7.9 (0.2) &  6.8 (0.2) \\
      \bottomrule
    \end{tabular}
  }%
    \caption{Average Detection Delays (ADDs) of different methods with \textbf{noise} for change pattern ($p=100$).}
  \label{tab:p=100noise}
\end{table}

\subsection{Different Shifts of Threshold in Training vs. Testing}

The performance of the network can be strongly influenced by the magnitude of mean shifts—for example, larger shifts are easier to detect. Consequently, the choice of training shift may bias the network’s detection capability. To assess this effect, we vary both the training and testing shifts in our experiments, demonstrating that the Causal DQ can robustly handle mismatches between training and testing mean shifts $\delta_{\text{train}} = 0.5$ and $\delta_{\text{train}} = 1$ of data streams. In this section, we train the Causal DQ  using two training shifts. We then evaluate each trained model across a range of testing shifts $\delta_{test}$ to assess how different training and testing magnitudes affect performance. We fix the settings similar to Section~\ref{sec:Noise-free}, where $\mu_{c}=[\underbrace{1, \ldots, 1}_k, 0, \ldots, 0] \times \delta$ and $k = 5$ for $p =10$, $k = 10$ for $p=50$, and $k = 20$ for $p=100$. The number of sensors is assigned as $ m= 6$ for $p =10$, $m = 12$ for $p=50$, and $m = 22$ for $=100$ with time steps $t=200$. Meanwhile, we also train Non-Causal DQ networks under the same settings as baseline. 

The results in Table~\ref{tab:train_test} show that
Non-Causal DQ exhibits consistently higher detection delays across all training scenarios. In contrast, our proposed causal DQ consistently achieves low detection delays across all training shift values, indicating strong generalization ability and robustness to the choice of training shift magnitude

\begin{table}[H]
\centering

\resizebox{\textwidth}{!}{%
\begin{tabular}{@{}r cc cc cc @{\quad} cc cc cc@{}}
  \toprule
  & \multicolumn{6}{c}{\bf Causal DQ} & \multicolumn{6}{c}{\bf Non-Causal} \\
  \cmidrule(lr){2-7}\cmidrule(lr){8-13}
  $\delta_{\rm test}$
    & \multicolumn{2}{c}{$p=10$} & \multicolumn{2}{c}{$p=50$} & \multicolumn{2}{c}{$p=100$}
    & \multicolumn{2}{c}{$p=10$} & \multicolumn{2}{c}{$p=50$} & \multicolumn{2}{c}{$p=100$} \\
  \cmidrule(lr){2-3}\cmidrule(lr){4-5}\cmidrule(lr){6-7}
  \cmidrule(lr){8-9}\cmidrule(lr){10-11}\cmidrule(lr){12-13}
  & $\delta_{\text{train}} = 0.5$ & $\delta_{\text{train}} = 1$ & $\delta_{\text{train}} = 0.5$ & $\delta_{\text{train}} = 1$ & $\delta_{\text{train}} = 0.5$ & $\delta_{\text{train}} = 1$
  & $\delta_{\text{train}} = 0.5$ & $\delta_{\text{train}} = 1$ & $\delta_{\text{train}} = 0.5$ & $\delta_{\text{train}} = 1$ & $\delta_{\text{train}} = 0.5$ & $\delta_{\text{train}} = 1$ \\
  \midrule
  0.25 & \textbf{60.7} (3.2) & 64.4 (4.0) & 117.2 (8.3) & \textbf{110.5} (7.2) & 128.6 (4.9) & \textbf{124.3} (5.2)
       & 72.7 (4.0) & 71.3 (4.7) & 118.2 (9.0) & 115.4 (8.5) & 133.2 (8.2) & 127.4 (7.8) \\
  0.50 & \textbf{18.1} (2.8) & 19.5 (3.2) & \textbf{22.2} (2.8)  & 25.7 (1.3)  & 29.4 (4.7)  & \textbf{29.2} (3.3)
       & 27.1 (2.0) & 27.4 (2.3) & 30.4 (1.8)  & 29.1 (2.1)  & 35.8 (5.2)  & 33.2 (4.0) \\
  1.00 & \textbf{10.1} (1.6) & 10.8 (2.1) & 13.4 (0.2)  & \textbf{12.2} (0.3)  & \textbf{14.2} (0.3)  & 14.6 (0.3)
       & 14.0 (1.8) & 13.3 (2.4) & 13.3 (0.4)  & 13.8 (0.2)  & 17.1 (0.8)  & 17.2 (0.8) \\
  1.50 &  5.2 (0.4) &  \textbf{4.9} (0.7) &  \textbf{5.4} (0.1)  &  5.7 (0.4)  &  6.5 (0.3)  &  \textbf{6.4} (0.4)
       &  6.1 (0.1) &  6.3 (0.1) &  6.7 (0.4)  &  6.3 (0.6)  &  8.4 (0.3)  &  7.4 (0.2) \\
  2.00 &  4.8 (0.3) &  \textbf{4.2} (0.2) &  5.6 (0.4)  &  \textbf{5.1} (0.2)  &  \textbf{5.8} (0.2)  &  5.9 (0.4)
       &  6.4 (0.1) &  6.2 (0.2) &  5.9 (0.1)  &  5.2 (0.1)  &  6.2 (0.2)  &  6.2 (0.2) \\
  \bottomrule
\end{tabular}%
} 
\caption{Causal vs. Non-Causal DQ results across different \(p\) and \((\delta_{\rm train}, \delta_{\rm test})\) pairs}
\label{tab:train_test}
\end{table}

\subsection{Extreme Case}
In this section, we simulate two extreme scenarios to evaluate the Causal DQ’s ability to detect subtle mean shifts and amount of sensors in high-dimensional data streams, where $k = 3$, $m=3$ for $p=50$, and $k = 6$, $m=6$ for $p=100$ as  $\mu_{c}=[\underbrace{1, \ldots, 1}_k, 0, \ldots, 0] \times \delta$. We fix the time steps of samples as $t=300$ in this section. The results are reported in Table~\ref{tab: extremecase}.

All methods struggle to detect anomalies within the allotted time steps when the mean shift is slight (e.g. $\delta = 0.25$). Moreover, even under large shifts ($\delta = 2$), their ADDs increase markedly, highlighting these scenarios’ difficulty for partially observable sensors. However, our Causal DQ consistently achieves the smallest ADD in both challenging cases,which displays its ability to detect subtle mean shifts in high-dimensional data streams. For example, when $\delta=0.5, p=50$, and $m=3$, Causal DQ achieves an ADD of 72.3, which is faster than all other benchmark methods. The performance advantage of Causal DQ becomes even more pronounced as the mean shift increases (e.g., $\delta=1.5$ and $\delta=2$ ).

\begin{table}[H]
\centering

\resizebox{\textwidth}{!}{%
  \begin{tabular}{@{} c r r r r r @{\quad} c r r r r r @{}}
  \toprule
   & \multicolumn{5}{c}{$p=50$, $m=3$} & & \multicolumn{5}{c}{$p=100$, $m=6$} \\
  \cmidrule(lr){2-6} \cmidrule(lr){8-12}
  $\delta$ 
   & Causal DQ & Non-Causal & TSSRP & ASMP  & SDQ 
   & & Causal DQ & Non-Causal & TSSRP & ASMP  & SDQ  \\
  \midrule
  0.25 & \textbf{300.0} (0.0) & 300.0 (0.0) & 300.0 (0.0) & 300.0 (0.0)& 300.0 (0.0)
       & & \textbf{300.0} (0.0) & 300.0 (0.0) & 300.0 (0.0) & 300.0 (0.0) & 300.0 (0.0) \\
  0.50 & \textbf{72.3} (6.6) & 82.4 (7.8) & 104.3 (8.9)  & 83.9 (7.2)  & 97.1 (6.8)
       & & \textbf{79.3} (6.1) & 96.7 (3.5) & 145.2 (8.4) & 90.9 (5.7) & 100.3 (9.2) \\
  1.50 &  \textbf{25.8} (1.2) & 32.6 (1.3) &  38.2 (2.2)  & 39.0 (3.7) & 30.1 (2.0)
       & &  \textbf{58.1}(3.8) & 67.6 (5.3) &  60.7 (6.1) & 59.3 (8.2) & 65.2 (4.2) \\
  2.00 &  \textbf{16.2} (1.6) & 24.8 (1.8) &  22.3 (2.0) & 25.3 (1.1) & 27.6 (1.9)
       & &  \textbf{23.4} (1.6) & 27.7 (2.1) &  30.7 (1.7) & 29.4 (1.2) & 28.3 (2.3) \\
  \bottomrule
  \end{tabular}%
} 
\caption{Average Detection Delays (ADDs) of different methods for small amount of mean shifts.}
\label{tab: extremecase}
\end{table}

\subsection{Ablation Study}

During the training of Causal DQ, we employ causal discovery methods to estimate the causal relationships among the $m$ selected variables, then computing the corresponding causal statistic to construct the causal state  (see Section~\ref{sec:3.2causaldiscover} and Section~\ref{sec:causalSA}). A natural question arises: to what extent does the performance of causal discovery affect the overall effectiveness of Causal DQ? In this section, we will perform an ablation study.

To evaluate the reliability of the estimated causal graph, we compare it with the ground truth by counting the number of missing, extra, and reversed edges. This provides a basic validation for the accuracy of the causal statistic computation. 
We assess performance using Structural Hamming Distance (SHD), True Positive Rate (TPR), and False Discovery Rate (FDR) \citep{yoav1995controlling,tsamardinos2006max,glymour2019review, chen2024temporal}. SHD evaluates how much the inferred graph deviates from the true structure by counting missing, extra, and misdirected edges—lower values indicate better accuracy. TPR reflects the fraction of true edges correctly recovered, while FDR captures the proportion of incorrect edges among all predicted ones; high TPR and low FDR imply more reliable estimation of causal graph. We report the average of the 3 metrics (mSHD, mTPR, mFDR) in Table~\ref{tab:causal_metrics} over the final 50 epochs-after cumulative reward has converged-for both $p=10$ and $p=50$ with different noises associated with Section~\ref{sec:Noise-free} and Section~\ref{sec:diffNoise}.

Table~\ref{tab:causal_metrics} shows that when $p =10$, the estimated causal graph achieves high accuracy, with $\text{mTPR} \geq 0.6$ and most $\text{mFDR} \leq 0.3$. Even in the high-dimensional setting ($p = 50$), all three metrics remain within a reasonable range, meeting commonly accepted standards \citep{zheng2018dags, chen2024temporal}. This level of performance is notable given two compounding difficulties: the high dimensionality of the system and the fact that we only observe a subset of variables ($m \ll p$), which limits the available information for accurate causal discovery. Moreover, the causal graph in our simulated dataset is relatively dense, which makes it more challenging to accurately estimate the full set of causal relationships.

\begin{table}[H]
\centering

\begin{tabular}{l|ccc|ccc}
\toprule
\textbf{Noise ($\sigma$)} & \multicolumn{3}{c|}{$p = 10$, $m=6$} & \multicolumn{3}{c}{$p = 50$, $m=12$} \\
& mSHD & mTPR & mFDR & mSHD & mTPR & mFDR \\
\midrule
Free   & 5.7  &  0.65  &  0.21  &  33.2  &  0.49  &  0.42 \\
0.05   &  9.1  &  0.60 &  0.33 &  33.0  &  0.51  &  0.41 \\
0.10  &  5.2  &  0.67  &  0.20  &  38.0  &  0.45  &  0.48 \\
0.15   &  6.3  &  0.63 &  0.29 &  30.4  &  0.53  &  0.40\\
\bottomrule
\end{tabular}
\caption{Metrics of Causal Graph under Different Noise Levels and Dimensions}
\label{tab:causal_metrics}
\end{table}

To further investigate the relationship between causal discovery accuracy and Average Detection
Delay, we conducted experiments with the following setup: $p=50, m=12$, mean shift $\delta=1$, and noise level $\sigma=0.1$. The total time step is $T=200$. 
Notably, we deliberately constructed high-quality and low-quality causal estimation for the causal graph over the selected $m$ variables. The results are reported in Table~\ref{tab:3metmore}. 

In the low-quality causal estimation setting, we include an extreme case \textbf{where the causal discovery method completely fails to recover any true causal relationships} among variables (second row in Table~\ref{tab:3metmore}). As a result, mTPR is $0$, and mFDR is $\frac{0}{0}$ which is denoted as non-defined (N/A). We omit the mSHD value in this case (denoted as “–”), as counting structural differences is meaningless when no edges are recovered at all. Similarly, the last row of Table~\ref{tab:3metmore} represents an extreme case where all \textbf{true causal relationships among variables are perfectly recovered}. To construct this setting, we directly use the ground-truth causal graph from the simulated dataset, where mTPR is $1$, and both mSHD and  mFDR are $0$. 


Table~\ref{tab:3metmore} demonstrates that the performance of Causal DQ is closely tied to the quality of the estimated causal relationships. The first row presents the performance of the \textbf{non-causal version} under the same experimental setting, served as a baseline for comparison, where the three causal graph metrics are marked as “–” since no causal structure is provided.  As the accuracy of causal graph estimation improves, the detection of mean shifts becomes more effective. Specifically, when the causal graph is perfectly recovered, the ADD is $9$, which is significantly faster than in cases with lower estimation accuracy.
Importantly, if the causal relationships are completely wrong recovered, Causal DQ may fail to detect any mean shift at all, as the resulting \textbf{causal state} (Section~\ref{sec:causalSA}) becomes invalid and the training network may be entirely misaligned. This paper directly employs existing causal discovery methods to recover the causal graph among variables, as improving causal discovery algorithms is not the focus of this work. Therefore, different causal discovery methods may lead to varying levels of performance in the estimated causal graphs.

In conclusion, when the TPR exceeds 0.4 (as shown in Table~\ref{tab:causal_metrics} and Table~\ref{tab:3metmore}), Causal DQ achieves a lower ADD than the non-causal baseline. In contrast, when the TPR falls below 0.2, the performance of Causal DQ degrades significantly due to a large proportion of incorrectly estimated causal relationships.
Moreover, most existing causal discovery methods are able to achieve a TPR greater than 0.4, suggesting that Causal DQ is robust and can be effectively integrated with a wide range of causal discovery algorithms. 




\begin{table}[H]
\centering

\resizebox{\textwidth}{!}{
\begin{tabular}{lcccc}
\toprule
\textbf{Scenario} & \textbf{mSHD} & \textbf{mTPR} & \textbf{mFDR} & \textbf{ADD} \\
\midrule
Non-causal version & – & – & – & \textbf{14} \\
Completely fails to recover (no edges recovered)     & –   & 0.00 & N/A & \textbf{200} \\
Low-quality causal estimation            & 60.7 & 0.17 & 0.89 & \textbf{79} \\
Standard-quality causal estimation       & 35.0 & 0.51 & 0.46 & \textbf{12} \\
Perfectly recovered (ground-truth causal graph)   & 0.0  & 1.00 & 0.00 & \textbf{9} \\
\bottomrule
\end{tabular}}
\caption{Performance comparison of Causal DQ under different causal graph qualities, and baseline Non-causal version.  
Experiment setup: $p=50$, $m=12$, mean shift $\delta=1$, noise level $\sigma=0.1$, and $T=200$.}
\label{tab:3metmore}
\end{table}

\section{Real Case}\label{sec:realcase}
\subsection{Tennessee Eastman Process (TEP)}
To demonstrate the effectiveness of our proposed Causal DQ network, we evaluate its performance on the Tennessee Eastman Process (TEP), a well-known industrial benchmark problem for fault detection and diagnosis in chemical engineering. TEP simulates a complex chemical production process with interacting control loops, stochastic disturbances, and a range of process faults, making it an ideal testbed for evaluating change-point detection and anomaly diagnosis algorithms.

The datasets used in our experiments are provided by \cite{rieth2017additional}, which include both training and testing time series collected under normal and faulty operating conditions. Each dataset contains 55 columns, where columns 5 through 55 represent the fully observed process variables (e.g., temperature, pressure, flow rates), yielding a high-dimensional monitoring space with $p = 50$. The remaining columns encode control signals and fault labels.

\begin{figure}[H] 
\centering
  \includegraphics[width=1\textwidth]{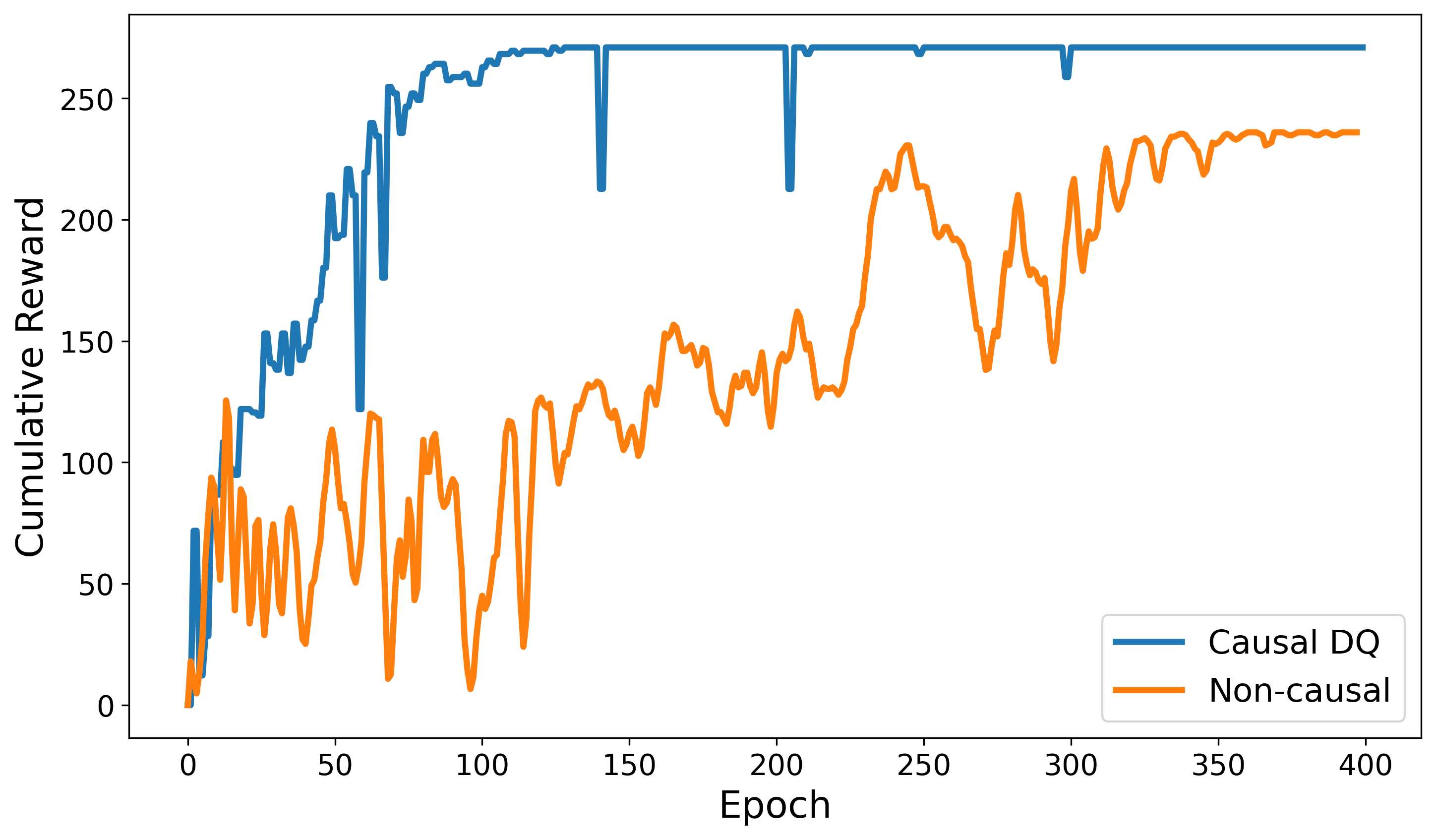} 
  \caption{Tennessee Eastman Process (TEP); Total Reward Comparison between Causal and Non-Causal DQ of real datasets as $p=50$. }              
  \label{fig:real_case}                 
\end{figure}

We use the first 500 samples (e.g., $t=500$) from the original training set, which contains no anomalies, for training the Causal DQ network. To train the Causal DQ network, we randomly select $10$ variables as mean shift data streams with the magnitude  $\delta = 1$. Similarly, we also train the Non-Causal $Q$-network under the same settings. The trained results are provided in Figure~\ref{fig:real_case}. On these real datasets, our Causal DQ consistently converges faster than the Non-causal version.

After training the Causal DQ network, we randomly sample $500$ testing observations ($t=500$, $p=50$) from the 480,000 in-control testing instances. We then inject a mean shift of magnitude $\delta$ into $m=5$ and $m=10$ randomly chosen data streams. 

The resulting ADDs are reported in Table~\ref{tab:real_case}. We can find that the Causal DQ outperforms other baselines even with a limited number of sensors and subtle mean shifts, which is consistent with the above results in Section~\ref{sec:Experiment}.

\begin{table}[H]
\centering

\resizebox{\textwidth}{!}{%
  \begin{tabular}{@{} c r r r r r @{\quad} c r r r r r @{}}
  \toprule
   & \multicolumn{5}{c}{m = 5} & & \multicolumn{5}{c}{m = 10} \\
  \cmidrule(lr){2-6} \cmidrule(lr){8-12}
  $\delta$ 
   & Causal DQ & Non-Causal & TSSRP & ASMP  & SDQ 
   & & Causal DQ & Non-Causal  & TSSRP & ASMP & SDQ \\
  \midrule

  0.25 & \textbf{171.4} (4.2) & 193.5 (5.0) & 170.2 (2.3)& 187.3 (2.8) & 190.4 (3.7)
       & & \textbf{88.2} (3.9) & 96.4 (3.8) & 98.7 (4.6) & 90.8 (3.9) & 92.3 (3.2) \\

  0.50 & \textbf{58.2} (3.1) & 67.8 (3.7) & 60.8 (2.0)& 76.7 (2.1) & 64.3 (3.2)
       & & \textbf{30.9} (2.7) & 39.4 (4.8) & 37.2 (5.2) & 40.3 (4.6) & 43.6 (3.7) \\
  1.00 &  \textbf{49.3} (1.5) & 54.6 (3.0)  &  16.3 (0.3) & 15.7 (0.2) &16.8 (0.7)
       & &  \textbf{12.9} (0.8) & 17.6 (1.3)&  18.8 (0.2)  & 19.5 (0.5) &18.1 (0.3) \\
 
  2.00 &  \textbf{20.1} (0.4) & 28.5 (0.8) &  27.3 (0.5) & 25.8 (0.6) & 24.6 (0.3)
       & &  \textbf{5.4} (0.3) & 6.8 (0.2) &  8.8 (0.4)  & 11.7 (0.2) & 10.8(0.4) \\
  \bottomrule
  \end{tabular}%
} 
\caption{Average Detection Delays (ADDs) of different methods for real datasets}
\label{tab:real_case}
\end{table}

\subsection{Solar Flare Detection (SFD)}

Our proposed \textbf{Causal DQ} is further evaluated on a real-world dataset—the Solar Flare dataset \cite{bigdata2020-flare-prediction}—to demonstrate its effectiveness in practical applications. Solar flares are solar events that can disrupt space operations and radio communications, posing risks to astronauts and technology-dependent systems. This dataset records relevant variables—such as the total magnitude of Lorentz force (TOTBSQ), total photospheric magnetic energy density (TOTPOT), and the mean value of the total field gradient (MEANGBT)—both before and during solar flare events. Samples are labeled as Q when no flare occurs, and as B, C, or M when flares are detected, corresponding to increasing levels of intensity from weak to strong. Each sample is observed as a time series from time step $1$ to $60$. In this case, we treat samples labeled as Q (no flare occurs)  as normal data streams. To demonstrate the capability of \textbf{Causal DQ} in handling challenging scenarios, we adopt samples labeled as B—representing the weakest occurrence of solar flares—as mean-shift anomalies in our implementation. Since weaker solar flares induce only subtle changes in the related variables, detecting such anomalies using partial sensors becomes significantly more challenging.

\paragraph{Training Dataset.} We selected Sample 80 (labeled as Q) and Sample 17 (labeled as B) to construct the training dataset. Specifically, we randomly extracted 30 rows from Sample 17 and inserted them into Sample 80, resulting in a new dataset with a total of 90 time steps. This dataset contains 60 time steps of normal data stream and 30 time steps exhibiting a mean shift. The dataset includes 33 variables, i.e., the dimensionality is $p = 33$ in this case. Furthermore, we set $m = 6$ and $m = 10$ to evaluate the Average Detection Delay under different numbers of sensors.

The test dataset is constructed using Sample 33 (labeled as Q) and Sample 28 (labeled as B), following the same procedure as used for the training dataset described above.

\begin{figure}[H] 
\centering
  \includegraphics[width=1\textwidth]{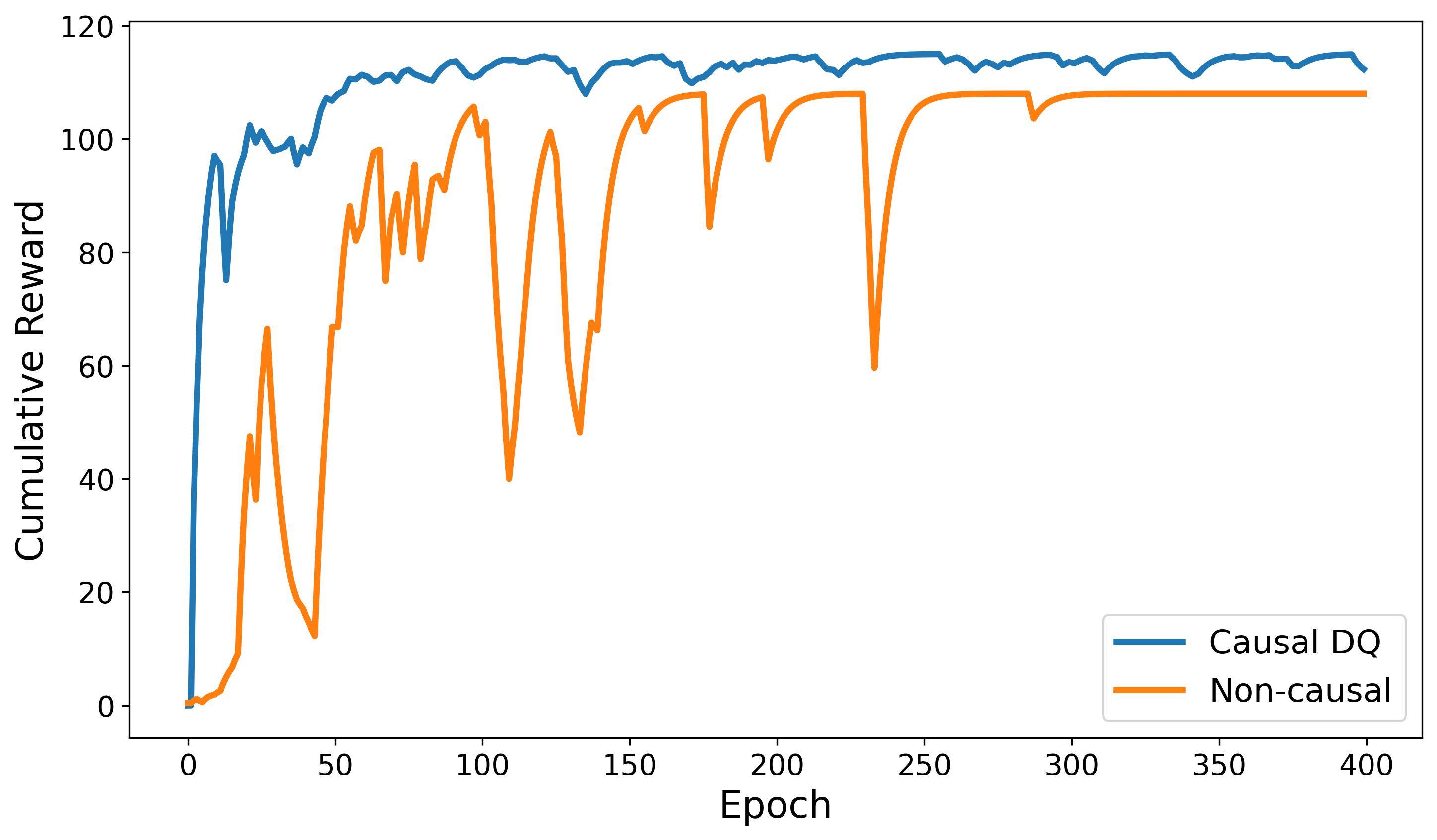} 
  \caption{Solar Flare Detection (SFD); Total Reward Comparison between Causal and Non-Causal DQ of real datasets as $p=33$. }              
  \label{fig:real_case2}                 
\end{figure}

The cumulative reward during training is shown in Figure~\ref{fig:real_case2}, which demonstrates that the proposed \textbf{Causal DQ} achieves a faster convergence rate compared to its non-causal counterpart.

Furthermore, the resulting Average Detection Delays (ADDs), reported in Table~\ref{tab:real_case2}, confirm that \textbf{Causal DQ} achieves the smallest detection delay in this real-world application.

\begin{table}[H]
\centering

\begin{tabular}{lcc}
\toprule
\textbf{Method} & \textbf{m = 6} & \textbf{m = 10} \\
\midrule
Causal DQ    & \textbf{12.8} (1.0) & \textbf{6.2} (0.8) \\
Non-Causal   & 15.6 (1.0)          & 8.1 (0.4) \\
TSSRP        & 14.1 (0.3)          & 7.9 (0.7) \\
ASMP         & 15.0 (0.5)          & 8.3 (1.0) \\
SDQ          & 18.2 (0.7)          & 10.6 (0.8) \\
\bottomrule
\end{tabular}
\caption{Solar Flare Detection (SFD); Average Detection Delays (ADDs) of different methods for real datasets}
\label{tab:real_case2}
\end{table}

\section{Conclusion}\label{sec:discuss}

We propose a novel framework, Causal DQ, for anomaly detection in partially observable sensor environments, which integrates a non-interventional causal approach with reinforcement learning. In particular, the non-interventional causal framework makes Causal DQ scalable and compatible with a wide range of RL methods. Causal DQ achieves faster convergence, tighter error bounds, and reduced anomaly detection time across various settings, demonstrating strong performance in large-scale, real-world partially observable environments. Both simulations and real-world applications demonstrate that Causal DQ achieves the shortest Average Detection Delay across various scenarios. Beyond anomaly detection, Causal DQ also shows promise in other fields such as robotics, computer vision, image analysis, and intelligent decision-making. We leave the exploration of these domains to future work.

\newpage
\appendix
\section{Appendix}
\subsection{Mathematical Proof}

\noindent \textbf{Proof of Lemma \ref{lem:lemma1}}

\begin{proof}
Let
$$
\varepsilon(s,a)
:= \bigl(\mathcal T_c^\pi Q\bigr)(s,a)
   - \bigl(\mathcal T_c^\pi Q'\bigr)(s,a).
$$
By definition of $\mathcal T_c^\pi$, the reward $r(s,a)$ and the causal entropy term 
$\tau\,\mathcal H_c(\pi(\cdot\mid s'))$ cancel when taking the difference, so
$$
\begin{aligned}
&\varepsilon(s,a)\\
&= \gamma\,
  \mathbb E_{s'}\Bigl[
    \sum_{a'}\pi(a'\mid s')\,Q(s',a') -\tau H_c\left(\pi\left(\cdot \mid s^{\prime}\right)\right)
    - \sum_{a'}\pi(a'\mid s')\,Q'(s',a')+\tau H_c\left(\pi\left(\cdot \mid s^{\prime}\right)\right)
  \Bigr]\\
&= \gamma\,
  \mathbb E_{s'}\Bigl[
    \sum_{a'}\pi(a'\mid s')\bigl(Q(s',a') - Q'(s',a')\bigr)
  \Bigr].
\end{aligned}
$$
This implies that,
$$
\begin{aligned}
&\bigl|\varepsilon(s,a)\bigr|\\
&\le \gamma\,
  \mathbb E_{s'}\Bigl[
    \sum_{a'}\pi(a'\mid s')\,
    \bigl|Q(s',a') - Q'(s',a')\bigr|
  \Bigr]\\
&\le \gamma\,
  \mathbb E_{s'}\Bigl[
    \max_{a'}\bigl|Q(s',a') - Q'(s',a')\bigr|
  \Bigr]
\le \gamma\,\|Q - Q'\|_\infty.
\end{aligned}
$$
Since this bound holds for all $(s,a)$,
$$
\bigl\|\mathcal T_c^\pi Q - \mathcal T_c^\pi Q'\bigr\|_\infty
= \sup_{s,a}\bigl|\varepsilon(s,a)\bigr|
\le \gamma\,\|Q - Q'\|_\infty
$$
which ensures the contraction of the \textbf{causal entropy-regularized} $Q$-function. 

\end{proof}

\noindent \textbf{Proof of Proposition \ref{prop:th1.2}}

\begin{proof}[Proof] 
As defined in Section~\ref{tab:causalEntropy}, the causal entropy is generally expressed as: 

$$\mathcal{H}_c(\pi(\cdot \mid s))=-\sum_{a_i} \mathbf{C}_{a_i \rightarrow r \mid s} \pi\left(a_i \mid s\right) \log \pi\left(a_i \mid s\right)$$ 

Then, we consider it in the following form:

$$f(\pi)=\sum_{i=1}^n \mathbf{C}_i \pi_i \log \pi_i ,\quad \pi_i \geq 0, \quad \sum_{i=1}^n \pi_i=1$$

where $\pi_i=\pi\left(a_i \mid s\right)$ under policy distribution $\pi$.

Let's define $x = \pi$, $y = \pi^\prime$ and $z = \lambda x + (1 - \lambda) y$, where $\lambda \in[0,1]$. So we can have: 

$$\begin{aligned}
f(z)&=\sum_{i=1}^n \mathbf{C}_i z_i \log z_i\\
&=\sum_{i=1}^n \mathbf{C}_i\left[\lambda x_i+(1-\lambda) y_i\right] \log \left[\lambda x_i+(1-\lambda) y_i\right]
\end{aligned}$$ 

As $t\log(t)$ is convex function, we can derive that:

$$(\lambda x+(1-\lambda) y) \log (\lambda x+(1-\lambda) y) \leq \lambda x \log x+(1-\lambda) y \log y$$ which implies that 

$$\mathbf{C}_i\left(\lambda x_i+(1-\lambda) y_i\right) \log \left(\lambda x_i+(1-\lambda) y_i\right)\\
\leq \lambda \mathbf{C}_i x_i \log x_i+(1-\lambda) \mathbf{C}_i y_i \log y_i$$

Then, we can have 
$$\begin{aligned}
&\sum_{i=1}^n \mathbf{C}_i\left(\lambda x_i+(1-\lambda) y_i\right) \log \left(\lambda x_i+(1-\lambda) y_i\right)=\sum_i \mathbf{C}_i z_i \log z_i \\
&\leq \lambda \sum_i \mathbf{C}_i x_i \log x_i+(1-\lambda) \sum_i \mathbf{C}_i y_i \log y_i\\
&=\lambda f(x)+(1-\lambda) f(y)
\end{aligned}$$
Note:
$$\lambda f(x)+(1-\lambda) f(y)=\lambda \sum_{i=1}^n \mathbf{C}_i x_i \log x_i+(1-\lambda) \sum_{i=1}^n \mathbf{C}_i y_i \log y_i$$ which implies that $f(\pi)=\sum_{i=1}^n \mathbf{C}_i \pi_i \log \pi_i$ is convex.  

Therefore, we can conclude that negative causal entropy $-\mathcal{H}_c(\pi(\cdot \mid s))$ is convex, as $\mathcal{H}_c(\pi(\cdot \mid s))$ is concave (The negative of a convex function is a concave function).
\end{proof}

\noindent \textbf{Proof of Theorem \ref{thm:th1.1}}
\begin{proof}[Proof] 
As the \textit{Theorem 1} in \cite{adamczyk2024boosting}, the \textbf{Non-causal entropy-regularized} optimal value function $Q^*(s, a)$ is bounded as: 

$$r(s, a)+\gamma\left({\mathbb{E}} V\left(s^{\prime}\right)+\frac{\inf \Delta(s, a)}{1-\gamma}\right) \leq Q^*(s, a) \leq r(s, a)+\gamma\left({\mathbb{E}} V\left(s^{\prime}\right)+\frac{\sup \Delta(s, a)}{1-\gamma}\right)$$ where $$\Delta(s, a) = r(s, a)+\gamma {\mathbb{E}} V\left(s^{\prime}\right)-Q(s, a)$$ $$V(s) = 1 / \tau \log \mathbb{E}_{a \sim \pi} \exp \tau Q(s, a)$$ 

By log-sum-exp, we can have the lower bound as: 

$$\begin{aligned} 
&\sum_i \exp \left(\tau Q\left(s^{\prime}, a_i^{\prime}\right)\right) \geq \exp \left(\tau \max _{a^{\prime} \in  \mathcal{A}\left(s^{\prime}\right)} Q\left(s^{\prime}, a^{\prime}\right)\right)\\ &\Rightarrow  \log \sum_i \exp \left(\tau Q\left(s^{\prime}, a_i^{\prime}\right)\right) \geq \tau \max _{a^{\prime} \in  \mathcal{A}\left(s^{\prime}\right)} Q\left(s^{\prime}, a^{\prime}\right)
\end{aligned}
$$ which implies that 

$$\frac{1}{\tau}\log \sum_i \exp \left(\tau Q\left(s^{\prime}, a_i^{\prime}\right)\right) \geq \max _{a^{\prime} \in  \mathcal{A}\left(s^{\prime}\right)} Q\left(s^{\prime}, a^{\prime}\right)$$

Similarly, the upper bound of $V(s)$ can be derived as: 

$$\begin{aligned}
\sum_i \exp \left(\tau Q\left(s^{\prime}, a_i^{\prime}\right)\right) &\leq \sum_i\exp \left(\tau \max _{a_i^{\prime} \in  \mathcal{A}\left(s^{\prime}\right)} Q\left(s^{\prime}, a_i^{\prime}\right)\right)\\
\end{aligned}
$$ which implies that 

$$\begin{aligned}
\log\sum_i \exp \left(\tau Q\left(s^{\prime}, a_i^{\prime}\right)\right) &\leq \log\left(|\mathcal{A}| \exp \left(\tau \max _{a_i^{\prime} \in  \mathcal{A}\left(s^{\prime}\right)} Q\left(s^{\prime}, a_i^{\prime}\right)\right)\right)\\
&=\log\left(|\mathcal{A}|\right)+ \tau \max _{a_i^{\prime} \in  \mathcal{A}\left(s^{\prime}\right)} Q\left(s^{\prime}, a_i^{\prime}\right)
\end{aligned}
$$

Thus, we can derive the bound as:
$$\max _{a^{\prime} \in  \mathcal{A}\left(s^{\prime}\right)} Q\left(s^{\prime}, a^{\prime}\right) \leq V (s) = \frac{1}{\tau} \log \sum_{a^{\prime}} \exp \left(\tau Q\left(s^{\prime}, a^{\prime}\right)\right) \leq \max _{a^{\prime} \in  \mathcal{A}\left(s^{\prime}\right)} Q\left(s^{\prime}, a^{\prime}\right)+\frac{\log |\mathcal{A}|}{\tau}$$ which implies 

$$r(s,a) \leq \Delta(s, a) \leq r(s,a) + \frac{\log |\mathcal{A}|}{\tau}$$

Then, we represent the causal entropy $\mathcal{H}_c $ by $V(s)$ as following: 

$$\begin{aligned} 
&\mathcal{H}_c(\pi(\cdot \mid s))\\
& =-\sum_{a_i}\mathbf{C}_{a_i\rightarrow r\mid s} \pi(a_i \mid s) \log \pi(a_i \mid s) \\ 
& =-\sum_{a_i} \mathbf{C}_{a_i\rightarrow r\mid s}\pi(a_i \mid s)\left[\tau Q(s, a_i)-\log \sum_{a^{\prime}} \exp \left(\tau Q\left(s, a^{\prime}\right)\right)\right] \\ 
& =-\tau \sum_{a_i} \mathbf{C}_{a_i\rightarrow r\mid s}\pi(a_i \mid s) Q(s, a_i)+\log \sum_{a^{\prime}} \exp \left(\tau Q\left(s, a^{\prime}\right)\right) \sum_{a_i} \mathbf{C}_{a_i\rightarrow r\mid s} \pi(a_i \mid s) \\
& =-\tau \sum_{a_i} \mathbf{C}_{a_i \rightarrow r \mid s} \pi(a_i \mid s) Q(s, a_i)+\tau V(s) \sum_{a_i} \mathbf{C}_{a_i \rightarrow r \mid s} \pi(a_i \mid s)\end{aligned}$$ 

which implies that the causal $V(s)$ denoted $V_c(s)$ as: $$V_c(s)=\frac{1}{\tau w(s)}\left[\mathcal{H}_c(\pi(\cdot \mid s))+\tau \sum_{a_i} \mathbf{C}_{a_i \rightarrow r \mid s} \pi(a_i \mid s) Q(s, a_i)\right]$$ where $w(s) =\sum_{a_i} \mathbf{C}_{a_i \rightarrow r \mid s} \pi(a_i \mid s) $. 

As $ 0 \leq \mathcal{H}_c(\pi(\cdot \mid s)) \leq \log \sum_{a_i}\mathbf{C}_{a_i \rightarrow r \mid s} $, we can easily derive the bound for this  $V_c(s)$ as: 

$$\max _{a^{\prime} \in  \mathcal{A}\left(s^{\prime}\right)} Q\left(s^{\prime}, a^{\prime}\right) \leq V_c(s) \leq   \max _{a^{\prime} \in  \mathcal{A}\left(s^{\prime}\right)} Q\left(s^{\prime}, a^{\prime}\right)+\frac{\log \sum_{a_i} \mathbf{C}_{a_i \rightarrow r \mid s}}{\tau}$$ which implies that $$r(s,a) \leq \Delta(s, a) \leq r(s,a) + \frac{\log \sum_{a_i} \mathbf{C}_{a_i \rightarrow r \mid s}}{\tau}$$ 

Finally, we can conclude that: 

$\begin{aligned} & Q^*(s, a) \geq r(s, a)+\gamma \mathrm{E}\left[\max _{a^{\prime} \in  \mathcal{A}\left(s^{\prime}\right)} Q\left(s^{\prime}, a^{\prime}\right)\right], \\ & Q^*(s, a) \leq r(s, a)+\gamma\mathrm{E}\left[\max _{a^{\prime} \in \mathcal{A}\left(s^{\prime}\right)} Q\left(s^{\prime}, a^{\prime}\right)\right]+\frac{\gamma}{1-\gamma} \frac{\log \sum_{a_i} \mathbf{C}_{a_i \rightarrow r \mid s}}{\tau} \end{aligned}$

\end{proof}

\noindent \textbf{Proof of Corollary \ref{tab:Corollary 1}}

\begin{proof}[Proof]

Let's define the $V(s)$ at the step $t$ as $V_t$ which can be expressed as: $V_t=1 / \tau \log \mathbb{E}_{a \sim \pi} \exp \tau Q_t(s, a)$, where $Q_t(s, a)=r_t(s, a)+\gamma \mathbb{E}_{s^{\prime}}\left[V_{t-1}\left(s^{\prime}\right)\right]$. 

We derive our proof using a similar approach to that of \cite{geist2019theory}, which established error bounds associated with the convergence of \textbf{Non-causal entropy-regularized} methods:

$$\left\|Q_t-Q^*\right\|_{\infty} \leq \gamma^t\left\|V_0-V^*\right\|_{\infty}+\frac{\gamma \ln |\mathcal{A}|}{1-\gamma}$$ where  $V^* = 1 / \tau \log \mathbb{E}_{a \sim \pi} \exp \tau Q^*(s, a)$.

Firstly, we can know  $$\max _{a^{\prime} \in  \mathcal{A}\left(s^{\prime}\right)} Q\left(s^{\prime}, a^{\prime}\right) \leq V_c(s) \leq   \max _{a^{\prime} \in  \mathcal{A}\left(s^{\prime}\right)} Q\left(s^{\prime}, a^{\prime}\right)+\frac{\log \sum_{a_i} \mathbf{C}_{a_i \rightarrow r \mid s}}{\tau}$$from Theorem~\ref{thm:th1.1}, which implies that $$\left\|V-V^*_c\right\|_{\infty} \leq\left\|Q-Q^*\right\|_{\infty}+ \frac{\log \sum_{a_i} \mathbf{C}_{a_i \rightarrow r \mid s}}{\tau}$$ for any $Q$-function.

As $\left\|V_t-V^*_c\right\|_{\infty} \leq \gamma\left\|V_{t-1}-V^*_c\right\|_{\infty}$ , this error bound can be expressed in an inductive form.: 

$$\begin{aligned}
&\left\|V_{t-1} - V^*_c\right\|_{\infty}\\
&\leq \gamma^{t-1} \left\|V_0 - V^*_c\right\|_{\infty} + \sum_{k=0}^{t-2} \gamma^k \frac{\log \left( \sum_{a_i} \mathbf{C}_{a_i \rightarrow r \mid s} \right)}{\tau}  \\
&= \gamma^{t-1} \left\|V_0 - V^*_c\right\|_{\infty} + \frac{\log \left( \sum_{a_i} \mathbf{C}_{a_i \rightarrow r \mid s} \right)}{\tau} \cdot \frac{1 - \gamma^{t-1}}{1 - \gamma}
\end{aligned}$$

Finally, we can obtain: $$
\begin{aligned}
&\left\|Q_t-Q^*\right\|_{\infty}\\
&\leq \gamma\left\|V_{t-1}-V^*_c\right\|_{\infty}\\
&\leq \gamma^t\left\|V_{0}-V^*_c\right\|_{\infty}+\frac{\gamma}{1-\gamma}\frac{\log \sum_{a_i} \mathbf{C}_{a_i \rightarrow r \mid s}}{\tau}
\end{aligned}$$
\end{proof}

\noindent \textbf{Proof of Corollary \ref{coro:corallary2}}

\begin{proof}[Proof] We first construct the inequality for $V_t(s)=1 / \tau \log \mathbb{E}_{a \sim \pi} \exp \tau Q_t(s, a)$ and $V_c^*$: 

$$\begin{aligned}
&\left|V_t(s)-V_c^*(s)\right| \\
& =\left|V_t(s)-\max _a Q_t(s, a)+\max _a Q_t(s, a)-V^*(s)\right| \\
& \leq\left|V_t(s)-\max _a Q_t(s, a)\right|+\left|\max _a Q_t(s, a)-V^*(s)\right|\end{aligned}$$

Then, we simply release the bound of $V_t(s)$ from 

$$\max _{a \in \mathcal{A}\left(s\right)} Q_t\left(s, a\right) \leq V_t(s) \leq \max _{a^{\prime} \in \mathcal{A}\left(s\right)} Q_t\left(s, a\right)+\frac{\log \sum_{a_i} \mathbf{C}_{a_i \rightarrow r \mid s}}{\tau}$$

to 

$$0 \leq \min _{a \in \mathcal{A}\left(s\right)} Q_t\left(s, a\right) \leq V_t(s) $$

We then can derive: 

$$\left|V_t(s)-\max _{a\in \mathcal{A}\left(s\right)} Q_t\left(s, a\right)\right| \leq \max _{a \in \mathcal{A}\left(s\right)} Q_t\left(s, a\right)-\min _{a \in \mathcal{A}\left(s\right)} Q_t\left(s, a\right)$$. 

which further implies that 
$$\begin{aligned}
&\max _a Q_t(s, a)-\min _a Q_t(s, a) \\&\leq \max _a r_t(s, a)-\min _a r_t(s, a)+\gamma\left(\max _{s^{\prime}} V_{t-1}\left(s^{\prime}\right)-\min _{s^{\prime}} V_{t-1}\left(s^{\prime}\right)\right)
\end{aligned}$$ as $Q_t(s, a)=r_t(s, a)+\gamma \mathbb{E}_{s^{\prime}}\left[V_{t-1}\left(s^{\prime}\right)\right]$. 

Notably, in Section~\ref{tab:reward}, the reward is determined by the causal reward mask $M$  which is  $r_{t} = g(M^{s \rightarrow r} \odot s_t, M^{a \rightarrow r} \odot a_t)$ otherwise negative constant. Thus, we can simply have $|r_t(s, a)| \leq g$ then $$\begin{aligned}
&\max _a Q_t(s, a)-\min _a Q_t(s, a) \\&\leq 2\cdot g+\gamma\left(\max _{s^{\prime}} V_{t-1}\left(s^{\prime}\right)-\min _{s^{\prime}} V_{t-1}\left(s^{\prime}\right)\right)\\
&\leq 2\cdot g \cdot (1+\gamma) + \gamma^2 \left(\max _{s^{\prime}} V_{t-2}\left(s^{\prime}\right)-\min _{s^{\prime}} V_{t-2}\left(s^{\prime}\right)\right)\\
&{\ldots}\\
& \leq \gamma^t \left(\max _{s^{\prime}} V_0\left(s^{\prime}\right)-\min _{s^{\prime}} V_0\left(s^{\prime}\right)\right)+2 \cdot g \cdot \sum_{k=0}^{t-1} \gamma^k \\
& =\gamma^t \left(\max _{s^{\prime}} V_0\left(s^{\prime}\right)-\min _{s^{\prime}} V_0\left(s^{\prime}\right)\right)+2\cdot g \cdot \frac{1-\gamma^t}{1-\gamma}
\end{aligned}$$ which implies that $$\begin{aligned}
&\left|V_t(s)-V_c^*(s)\right| \\
& \leq\left|V_t(s)-\max _a Q_t(s, a)\right|+\left|\max _a Q_t(s, a)-V^*(s)\right|\\
&\leq \gamma^t \left(\max _{s^{\prime}} V_0\left(s^{\prime}\right)-\min _{s^{\prime}} V_0\left(s^{\prime}\right)\right)+2\cdot g \cdot \frac{1-\gamma^t}{1-\gamma} + \gamma^t\left\|V_0-V^*\right\|_{\infty}
\end{aligned}$$

Thus, when $t \rightarrow \infty$, we can conclude that $$\left|V_t(s)-V_c^*(s)\right| \leq \frac{2 \cdot g}{1-\gamma} $$ which implies that  

$$\lim _{t \rightarrow \infty}\left\|Q_t-Q^*\right\|_{\infty} \leq \lim _{t \rightarrow \infty}\frac{1}{1-\gamma}\left\|V_t(s)-V_c(s)^* \right\|_{\infty} \leq \frac{2 \cdot g}{(1-\gamma)^2}$$ 

Finally, we already derived that $\lim _{t \rightarrow \infty}\left\|Q_t-Q^*\right\|_{\infty} \leq \frac{\gamma}{1-\gamma}\frac{\log \sum_{a_i} \mathbf{C}_{a_i \rightarrow r \mid s}}{\tau} $, and then we can finally conclude that: 

$$\lim _{t \rightarrow \infty}\left\|Q_t-Q^*\right\|_{\infty} \leq \min \left\{\frac{\gamma\log \sum_{a_i} \mathbf{C}_{a_i \rightarrow r \mid s}}{\tau(1-\gamma)}, \frac{2 g}{(1-\gamma)^2}\right\}$$ 

\end{proof}

\noindent \textbf{Proof of Lemma \ref{lem:lemma2}}
\begin{proof}[Proof]

According to Lemma \ref{lem:lemma1}, to guarantee that $\left\|Q_t-Q^*\right\|_{\infty} \leq \varepsilon$, it suffices to ensure that the exponentially decaying term satisfies

$$
\gamma^t\left\|V_0-V_c^*\right\|_{\infty} \leq \varepsilon-\frac{\gamma}{1-\gamma} \cdot \frac{\log \sum_{a_i} \mathbf{C}_{a_i \rightarrow r \mid s}}{\tau}
$$ where the right-hand side is positive associated with $$
\varepsilon>\frac{\gamma}{1-\gamma} \cdot \frac{\log \sum_{a_i} \mathbf{C}_{a_i \rightarrow r \mid s}}{\tau}
$$ which further implies that: 

$$\gamma^t \leq \frac{\varepsilon-\frac{\gamma}{1-\gamma} \cdot \frac{\log \sum_{a_i} \mathbf{C}_{a_i \rightarrow r \mid s}}{\tau}}{\left\|V_0-V_c^*\right\|_{\infty}} \Rightarrow t \geq \frac{\log \left(\frac{\varepsilon-\frac{\gamma}{1-\gamma} \cdot \frac{\log \sum_{a_i} \mathbf{C}_{a_i \rightarrow r \mid s}}{\tau}}{\left\|V_0-V_c^*\right\|_{\infty}}\right)}{\log \gamma}$$

Thus, we can finally conclude that the time steps $T$ required for $Q_t$ to approach $Q^*$ within a precision $\varepsilon$ (i.e., $\left\|Q_t-Q^*\right\|_{\infty} \leq \varepsilon$ and $\varepsilon-\frac{\gamma}{1-\gamma} \cdot \frac{\log \sum a_i \mathbf{C}_{a_i \rightarrow r \mid s}}{\tau} \rightarrow 0^{+}$ ) is asymptotically bounded by:

$$T = \mathcal{O}\left( \log\left( \frac{1}{\varepsilon - \frac{\gamma}{1 - \gamma} \cdot \frac{\log \sum_{a_i} \mathbf{C}_{a_i \rightarrow r \mid s}}{\tau}} \right) \right)$$

\end{proof}

\noindent \textbf{Proof of Theorem \ref{thm:ExpectError bounds}}

\begin{proof}[Proof]
\cite{lee2023discrete} and \cite{jeong2024finite} modeled the update rule of soft $Q$-learning at time step $t$ under stochastic noise $w_t$ as:
$$
Q_{t+1} = Q_t + \alpha \cdot (\mathcal{T}_t Q_k - Q_t + w_t)
$$
and further analyzed it using a non-linear switching system, yielding the following equivalent form:
$$
Q_{t+1} = Q_t + \alpha \left( D R + \gamma D P \Pi_{Q_t}^{\max} Q_t - D Q_t + w_t \right)
$$
where:
\begin{itemize}
  \item $D \in \mathbb{R}^{|\mathcal{S}||\mathcal{A}| \times |\mathcal{S}||\mathcal{A}|}$ is a diagonal matrix, where each diagonal element $d(s,a)$ represents the probability of sampling action $a \in \mathcal{A}$ at state $s \in \mathcal{S}$.
  \item $D$ can be expressed in a block-diagonal form over actions:
  $$
  D = \operatorname{diag}\left( D_a(1), \ldots, D_a(|\mathcal{S}|) \right), \quad \text{for each } a \in \mathcal{A}
  $$
  \item $P$ denotes the transition probability matrix.
  \item $R$ is the reward vector or matrix.
  \item $\Pi_{Q_t}^{\max} Q_t$ represents the operator applied to $Q_t$ to maximum it over actions at each state.
\end{itemize}

In this paper, we include the \textbf{causal entropy} $\mathcal{H}_c$ as regularization term. Notably, our $Q$-value function is not represented in a tabular form, but rather approximated by a neural network conditioned on the action space $\mathcal{A}$. As a result, the output of the Q -network is a vector whose dimension matches the cardinality of the action space, i.e., $|\mathcal{A}|$.
Furthermore, the Q-values are randomly sampled from the replay buffer, rather than through explicit enumeration of transition probabilities as in tabular methods. Thus, we replace the original transition probability matrix $D$ with a sampling-based matrix $\Omega \in$ $\mathbb{R}^{|\mathcal{A}| \times|\mathcal{A}|}$, constructed from the replay buffer. Here, $\Omega$ is a diagonal matrix, where each diagonal element $\omega \in(0,1)$ represents the empirical sampling weight associated with a particular state-action pair, which also reshapes the aggregated probability $P$ as $P\in$ $\mathbb{R}^{|\mathcal{A}| \times|S|}$

Therefore, we can represent the updating of our \textbf{Causal Q-network} as: 

$$
Q_{t+1} = Q_t + \alpha \left( \Omega R + \gamma \Omega P \Pi_{Q_t}^{\max} Q_t + \tau \Omega\mathcal{H}_c(Q_t) - \Omega Q_t + w_t \right)
$$

Here, the Bellman equation is: 

$$\left(\gamma \Omega P \Pi_{Q^*}-\Omega\right) Q^*+\Omega R-\tau D \mathcal{H}_c(Q^*)=0$$ which implies that 

$$\begin{aligned}
& Q_{t+1}\\
& =Q_t+\alpha\left[\Omega R+\gamma \Omega P \Pi_{Q_t} Q_t-\tau \Omega \mathcal{H}_c\left(Q_t\right)-\Omega Q_t+w_t\right] \\
&=Q_t+\alpha\left[\Omega Q^*-\gamma \Omega P \Pi_{Q^*} Q^*+\tau \Omega \mathcal{H}_c(Q^*)+\gamma \Omega P \Pi_{Q_t} Q_t-\tau \Omega \mathcal{H}_c(Q_t)-\Omega Q_t+w_t\right]
\end{aligned}$$

We then express it as:

$$Q_{t+1}-Q^*=A_{Q_t}\left(Q_t-Q^*\right)+b_{Q_t}+\alpha w_t$$ where 

$$\begin{gathered}
A_{Q_t}:=I+\alpha\left(\gamma \Omega P \Pi_{Q_t}-\Omega\right) \\\\
b_{Q_t}:=\alpha\left[\tau \Omega\left(\mathcal{H}_c(Q^*)-\mathcal{H}_c(Q_t)\right)+\gamma \Omega P\left(\Pi_{Q_t}-\Pi_{Q^*}\right) Q^*\right]
\end{gathered}$$

As the \cite{jeong2024finite}, we can have lower and upper comparison system donated as $Q_t^L \leq Q_t\leq Q_t^U$ :

$$\begin{aligned}
&\left(Q_{t+1}^L-Q^*\right)= A_{Q^*}\left(Q_t^L-Q^*\right) 
 +\alpha w_t-\alpha \gamma \Omega P \frac{\log \sum_{a_i} \mathbf{C}_{a_i \rightarrow r \mid s}}{\tau}\cdot \mathbf{1} \\
&\left(Q_{t+1}^U-Q^*\right)=A_{Q_t}\left(Q_t^U-Q^*\right)+\alpha w_t
\end{aligned}$$ which implies that: 

$$\begin{aligned}
& \left(Q_{t+1}^U-Q_{t+1}^L\right)\\
&= A_{Q_t}\left(Q_t^U-Q_t^L\right) +(A_{Q_t}-A_{Q^*})\left(Q_t^L-Q^*\right)+\alpha \gamma \Omega P \frac{\log \sum_{a_i} \mathbf{C}_{a_i \rightarrow r \mid s}}{\tau}\cdot \mathbf{1}
\end{aligned}$$

The norm of this equation can gives out: 

$$\begin{aligned}
& \left\|Q_{t+1}^U-Q_{t+1}^L\right\|_{\infty} \\
\leq & \left\|A_{{Q_t}}\right\|_{\infty}\left\|Q_t^U-Q_t^L\right\|_{\infty}+\left\|A_{Q_t}-A_{Q^*}\right\|_{\infty}\left\|Q_t^L-Q^*\right\|_{\infty} \\
& +\left\|\alpha \gamma \Omega P \frac{\log \sum_{a_i} \mathbf{C}_{a_i \rightarrow r \mid s}}{\tau}\right\|_{\infty} \\
\leq & \rho\left\|Q_t^U-Q_t^L\right\|_{\infty} \\
& +2 \alpha \gamma \omega_{\max }\left\|Q_t^L-Q^*\right\|_{\infty}+\frac{\alpha}{\tau} \log \sum_{a_i} \mathbf{C}_{a_i \rightarrow r \mid s}
\end{aligned}$$ where $\rho$ is exponential decay rate which is expressed as: $\rho:=1-\alpha \omega_{\min }(1-\gamma)$ \cite{lee2023discrete}. Meanwhile, $\left\|A_{Q_t}-A_{Q^*}\right\|_{\infty} \leq \alpha \gamma \omega_{\max }\left\|P\left(\Pi_{Q_t}^{\max }-\Pi_{Q^*}^{\max }\right)\right\|_{\infty} \leq 2 \alpha \gamma d_{\max }$ \citep{jeong2024finite}.

As $\left\|Q_t^L-Q^*\right\|_{\infty} \leq\left\|Q_t^L-Q^*\right\|_2$, we can consider the  $\left\|Q_t^L-Q^*\right\|_2$ as an upper bound for $\left\|Q_t^L-Q^*\right\|_{\infty}$, which is given as:

$$\begin{aligned}
& \mathbb{E}\left[\left\|Q_t^L-Q^*\right\|_2\right] \\
\leq & \underbrace{\sqrt{\mathbb{E}\left[\left\|A_{Q^*}^t\left(Q_0^L-Q^*\right)+\alpha \sum_{i=0}^{t-1} A_{Q^*}^{t-i-1} w_i\right\|_2^2\right]}}_{\mathbb{A}}  +\underbrace{\mathbb{E}\left[\left\|\alpha \gamma \sum_{i=0}^{t-1} A_{Q^*}^{t-i-1} \Omega P \frac{\log \sum_{a_i} \mathbf{C}_{a_i \rightarrow r \mid s}}{\tau}\cdot \mathbf{1} \right\|_2\right]}_\mathbb{B} 
\end{aligned}$$ by doing recursively. 

The term $\mathbb{A}$ is exactly the same as the expression in Theorem IV.1 of \cite{jeong2024finite} which has the upper bound as: $\left(|\mathcal{A}|\left\|Q_0^L-Q^*\right\|_2^2 \rho^{2 t}+\frac{6 \alpha| \mathcal{A}|}{\omega_{\min }(1-\gamma)^3}\right)^{\frac{1}{2}}$, while the term $\mathbb{B}$ corresponds to our proposed \textbf{causal component}. For term $\mathbb{B}$, we can obtain: 

$$\begin{aligned}
\left\|\alpha \gamma \sum_{i=0}^{t-1} A_{Q^*}^{t-i-1} \Omega P \frac{\log \sum_{a_i} \mathbf{C}_{a_i \rightarrow r \mid s}}{\tau}\cdot \mathbf{1} \right\|_2 & \leq \alpha \gamma \cdot\frac{\log \sum_{a_i} \mathbf{C}_{a_i \rightarrow r \mid s}}{\tau} \cdot \sum_{i=0}^{t-1}\left\|A_{Q^*}^{t-i-1} \Omega P \cdot \mathbf{1}\right\|_2 \\
& \leq \alpha \gamma \cdot \frac{\log \sum_{a_i} \mathbf{C}_{a_i \rightarrow r \mid s}}{\tau} \cdot \sum_{i=0}^{t-1} \rho^{t-i-1} \cdot \omega_{\max } \sqrt{|\mathcal{A}|}
\end{aligned}$$ where  $\left\|A_{Q^*}^{t-i-1}\right\|_{\infty} \leq \rho^{t-i-1}$ (See details in Lemma 3.3 of \cite{lee2023discrete}) and 
$\|\Omega P \cdot \mathbf{1}\|_2 \leq d_{\max } \cdot\|\mathbf{1}\|_2=\omega_{\max } \cdot \sqrt{|\mathcal{A}|}$.
Therefore, we can have: 

$$\begin{aligned}
\mathbb{E}\left[\left\|Q_t^L-Q^*\right\|_2\right]
&\leq \left(|\mathcal{A}|\left\|Q_0^L-Q^*\right\|_2^2 \rho^{2 t}+\frac{6 \alpha|\mathcal{A}|}{\omega_{\min }(1-\gamma)^3}\right)^{\frac{1}{2}}\\
&+ \alpha \gamma \cdot \frac{\log \sum_{a_i} \mathbf{C}_{a_i \rightarrow r \mid s}}{\tau} \cdot \sum_{i=0}^{t-1} \rho^{t-i-1} \cdot \omega_{\max } \sqrt{|\mathcal{A}|}
\end{aligned}$$

As $\left\|Q^*\right\|_{\infty} \leq \inf _{a_i \in \mathcal{A}}\frac{1}{1-\gamma}\left(1+\frac{\gamma}{1-\gamma} \cdot \frac{1}{\tau} \log \sum_{a_i} \mathbf{C}_{a_i \rightarrow r \mid s}\right) = \frac{\gamma}{1-\gamma}$, we can have:
$$\begin{aligned}
\left\|Q_0^L-Q^*\right\|_2 &\leq |\mathcal{A}|^{\frac{1}{2}} \left(1+\frac{1}{1-\gamma}\right)\\
&=|\mathcal{A}|^{\frac{1}{2}} \left(\frac{2}{1-\gamma}\right)
\end{aligned}$$ which finally gives out: 

$$\begin{aligned}
\mathbb{E}\left[\left\|Q_t^L-Q^*\right\|_2\right] & \leq\left(|\mathcal{A}|^{\frac{2}{3}} \left(\frac{2}{1-\gamma}\right)^2\rho^{2 t}+\frac{6 \alpha|\mathcal{A}|}{\omega_{\min }(1-\gamma)^3}\right)^{\frac{1}{2}} \\
& +\alpha \gamma \cdot \frac{\log \sum_{a_i} \mathbf{C}_{a_i \rightarrow r \mid s}}{\tau} \cdot \sum_{i=0}^{t-1} \rho^{t-i-1} \cdot \omega_{\max } \sqrt{|\mathcal{A}|}
\end{aligned}$$

Moreover, it is obvious that:  $Q_t^L -Q^* \leq Q_t-Q^*\leq Q_t^U-Q^*$. Then, for deriving the finite time step bound, we can have this inequality: 

$$\begin{aligned}
& \mathbb{E}\left[\left\|Q_t-Q^*\right\|_{\infty}\right] \\
= & \mathbb{E}\left[\left\|Q_t-Q_t^L+Q_t^L-Q^*\right\|_{\infty}\right] \\
\leq & \mathbb{E}\left[\left\|Q_t-Q_t^L\right\|_{\infty}\right]+\mathbb{E}\left[\left\|Q_t^L-Q^*\right\|_{\infty}\right] \\
\leq & \mathbb{E}\left[\left\|Q_t^U-Q_t^L\right\|_{\infty}\right]+\mathbb{E}\left[\left\|Q_t^L-Q^*\right\|_{\infty}\right]
\end{aligned}$$

The upper bound of $\mathbb{E}\left[\left|Q_t^U - Q_t^L\right|{\infty}\right]$ is derived in Theorem IV.2 as the \textbf{non-causal} version by applying Theorem V.1 from \cite{jeong2024finite}. In contrast, our proposed \textbf{causal entropy} modifies the Boltzmann operator by incorporating a causal mask, replacing the standard action space $|\mathcal{A}|$ with the masked sum $\sum_{a_i} \mathbf{C}_{a_i \rightarrow r \mid s}$. Therefore, the bound on $\mathbb{E}\left[\left\|Q_t^U - Q_t^L\right\|_{\infty}\right]$ under the \textbf{causal} setting can be expressed as: 

$$\begin{aligned}
& \mathbb{E}\left[\left\|Q_t^U-Q_t^L\right\|_{\infty}\right] \\
\leq & 2 \alpha \gamma \omega_{\max }|\mathcal{A}|^{\frac{1}{2}}\left\|Q_0^L-Q^*\right\|_2 t \rho^{t-1}+\frac{2 \sqrt{6} \alpha^{\frac{1}{2}} \gamma \omega_{\max }|\mathcal{A}|^{\frac{1}{2}}}{\omega_{\min }^{\frac{3}{2}}(1-\gamma)^{\frac{5}{2}}} \\
& +\frac{2 \gamma^2 \omega_{\max }^2 \log \sum_{a_i} \mathbf{C}_{a_i \rightarrow r \mid s}|\mathcal{A}|^{\frac{1}{2}}}{\tau \omega_{\min }^2(1-\gamma)^2}+\frac{\log \sum_{a_i} \mathbf{C}_{a_i \rightarrow r \mid s}}{\tau \omega_{\min }(1-\gamma)}\\
&= 2 \alpha \gamma \omega_{\max }|\mathcal{A}|\left(\frac{2}{1-\gamma}\right) t \rho^{t-1}+\frac{2 \sqrt{6} \alpha^{\frac{1}{2}} \gamma \omega_{\max }|\mathcal{A}|^{\frac{1}{2}}}{\omega_{\min }^{\frac{3}{2}}(1-\gamma)^{\frac{5}{2}}} \\
& +\frac{2 \gamma^2 \omega_{\max }^2 \log \sum_{a_i} \mathbf{C}_{a_i \rightarrow r \mid s}|\mathcal{A}|^{\frac{1}{2}}}{\tau \omega_{\min }^2(1-\gamma)^2}+\frac{\log \sum_{a_i} \mathbf{C}_{a_i \rightarrow r \mid s}}{\tau \omega_{\min }(1-\gamma)}
\end{aligned}$$

Therefore, we can have the finite time error bound for our proposed \textbf{causal entropy} $Q$-value as: 

$$\begin{aligned}
\mathbb{E}\left[\left\|Q_t-Q^*\right\|_{\infty}\right]
\leq 
& \frac{4 \alpha \gamma \omega_{\max }|\mathcal{A}|}{1-\gamma} \cdot t \cdot \rho^{t-1}+\frac{2 \sqrt{6} \alpha^{1 / 2} \gamma \omega_{\max }|\mathcal{A}|^{1 / 2}}{\omega_{\min }^{3 / 2}(1-\gamma)^{5 / 2}}\\
& +\frac{\log \sum_{a_i} \mathbf{C}_{a_i \rightarrow r \mid s}}{\tau}\left[\frac{2 \gamma^2 \omega_{\max }^2}{\omega_{\min }^2(1-\gamma)^2}+\frac{1}{\omega_{\min }(1-\gamma)}+\alpha \gamma \omega_{\max }|\mathcal{A}|^{1 / 2} \sum_{i=0}^{t-1} \rho^{t-i-1}\right] \\
& +\left(|\mathcal{A}|^{2 / 3} \cdot\left(\frac{2}{1-\gamma}\right)^2 \cdot \rho^{2 t}+\frac{6 \alpha|\mathcal{A}|}{\omega_{\min }(1-\gamma)^3}\right)^{1 / 2}
\end{aligned}$$
\end{proof}

\subsection{Table of Parameters}
We provide the details of the parameter settings for training Causal DQ in the simulation cases (Section~\ref{sec:Experiment}) in Table~\ref{tab:tuning}. The table includes the decay rate of causal entropy ($\alpha$), the learning rate (LR) for stochastic gradient descent (SGD), the discount factor ($\gamma$), the network architecture in terms of the number of units and layers (e.g., [256]$\times$4 denotes 4 layers with 256 units in each layer), the batch size, and the decay rate of exploration temperature ($\tau$).
\begin{table}[H]
\centering

\label{tab:qnetwork_tuning}
\begin{tabular}{lcccccc}
\toprule
\textbf{Setting} & $\boldsymbol{\alpha}$ &LR of SGD&$\boldsymbol{\gamma}$ & Hidden Units & Batch Size &$\tau$ \\
\midrule
$p=10$  & 0.05 & $5 \times 10^{-3}$ & 0.9  & [256]$\times 4$    & 32 & 0.65\\
$p=50$ & 0.1 & $1 \times 10^{-3}$ & 0.8  & [256]$\times 4$    & 64  & 0.75 \\
$p=100$ & 0.1 & $1 \times 10^{-3}$& 0.8  & [256]$\times 4$    & 64 & 0.9  \\
$p=10$ + noise & 0.05 & $5 \times 10^{-3}$ & 0.9  & [256]$\times 4$   & 64 & 0.65 \\ 
$p=50$ + noise  & 0.1 & $1 \times 10^{-3}$& 0.8  & [256]$\times 4$     & 128 & 0.75 \\
$p=100$ + noise & 0.1& $1 \times 10^{-3}$ & 0.8  & [256]$\times 4$ & 128& 0.9  \\
\bottomrule

\end{tabular}
\caption{Q-network Training and Tuning Results under Different Hyperparameter Settings; this table includes the decay rate of causal entropy ($\alpha$), the learning rate (LR) for stochastic gradient descent (SGD), the discount factor ($\gamma$), the network architecture in terms of the number of units and layers (e.g., [256]$\times$4 denotes 4 layers with 256 units in each layer), the batch size, and the decay rate of exploration temperature ($\tau$)}
\label{tab:tuning}
\end{table}

\bibliography{mybib}

\end{document}